\documentclass[a4paper,fleqn]{cas-sc}

\usepackage[numbers]{natbib}
\usepackage{amsmath,amssymb,amsfonts}
\usepackage{algorithmic}
\usepackage{algorithm}
\usepackage{array}
\usepackage{textcomp}
\usepackage{stfloats}
\usepackage{url}
\usepackage{verbatim}
\usepackage{graphicx}
\usepackage{multirow}
\usepackage{subcaption}
\usepackage{pifont}
\newcommand{\resultpm}[2]{#1\scriptsize\,$\pm$\,#2\normalsize}
\newcommand{\resultpmbf}[2]{\textbf{#1}\scriptsize\,$\pm$\,#2\normalsize}
\newcommand{\resultpmul}[2]{\underline{#1}\scriptsize\,$\pm$\,#2\normalsize}
\def\tsc#1{\csdef{#1}{\textsc{\lowercase{#1}}\xspace}}
\tsc{WGM}
\tsc{QE}
\tsc{EP}
\tsc{PMS}
\tsc{BEC}
\tsc{DE}

\begin{document}
\let\WriteBookmarks\relax
\def\floatpagepagefraction{1}
\def\textpagefraction{.001}
\shorttitle{A Multivariate Parameter-Efficient Fine-Tuning Framework for Few-shot Remaining Useful Life Prediction based on Cross-domain Time Series Representation Model}
\shortauthors{E. Fu et~al.}

\title [mode = title]{A Multivariate Parameter-Efficient Fine-Tuning Framework for Few-shot Remaining Useful Life Prediction based on Cross-domain Time Series Representation Model}                      
\tnotemark[1]

\tnotetext[1]{This work is supported by the National Natural Science Foundation of China (Grant 62273038), the Open Fund of Intelligent Control Laboratory (Grant 2024-ZKSYS-KF03-05) and the Ningbo Young Technological Innovation Leading Talent Project of China (Grant 2025QL047)}

\author[1]{En Fu}[style=chinese,
                  orcid=0009-0006-4837-6711]
\ead{fuen@xs.ustb.edu.cn}
\credit{Methodology, Software, Writing}

\author[2,3]{Yanyan Hu}[style=chinese, orcid=0000-0003-3578-7598]
\cormark[1]
\ead{huyanyan@ustb.edu.cn}
\credit{Methodology, Revision}


\author[4]{Zengwang Jin}[style=chinese, orcid=0000-0002-7491-9103]
\ead{Jin_zengwang@nwpu.edu.cn}
\credit{Revision}

\author[2,3]{Kaixiang Peng}[style=chinese, orcid=0000-0001-8314-3047]
\ead{kaixiang@ustb.edu.cn}
\credit{Revision}

\affiliation[1]{organization={School of Intelligence Science and Technology, University of Science and Technology Beijing},
                city={Beijing},
                postcode={100083}, 
                state={Beijing},
                country={China}}
\affiliation[2]{organization={School of Automation and Electrical Engineering, University of Science and Technology Beijing},
                city={Beijing},
                postcode={100083}, 
                state={Beijing},
                country={China}}
\affiliation[3]{organization={Key Laboratory of Knowledge Automation for Industrial Processes of the Ministry of Education, University of Science and Technology Beijing},
                city={Beijing},
                postcode={100083}, 
                state={Beijing},
                country={China}}
                

\affiliation[4]{organization={Institute of Embodied Intelligence, Anhui University},
                city={Hefei},
                postcode={230601}, 
                state={Anhui},
                country={China}}






\cortext[cor1]{Corresponding author}


\begin{abstract}
The application of data-driven remaining useful life (RUL) prediction has long been constrained by the availability of large amount of degradation data. Mainstream solutions like domain adaptation and meta-learning still rely on large amounts of historical degradation data from equipment identical or similar to the target, which imposes significant limitations in practical applications. This study investigates PEFT-MuTS, a Parameter-Efficient Fine-Tuning framework for few-shot RUL prediction, built on cross-domain pre-trained time-series models. Contrary to the view that knowledge transfer in RUL prediction can only occur within similar equipment, we demonstrate that substantial benefits are achieved through pre-training with large-scale cross-domain time series datasets. An independent feature tuning network and a meta-variable-based low rank multivariate fusion mechanism are developed to enable the pre-trained univariate time-series backbone model to fully exploit the multivariate relationships in degradation data for downstream RUL prediction. We introduce a zero-initialized regressor that stabilizes fine-tuning under few-shot conditions. Experiments on aero-engine and industrial bearing datasets demonstrate that our method achieves effective RUL prediction even when less than 1\% of samples of target equipment are used. It substantially outperforms conventional supervised and few-shot approaches while markedly reducing the data required to achieve high predictive accuracy. Code available at \url{https://github.com/fuen1590/PEFT-MuTS}.
\end{abstract}


\begin{highlights}
\item Cross-domain time-series priors enable accurate few-shot RUL prediction.
\item PEFT-MuTS efficiently adapts pretrained model to multivariate RUL prediction.
\item A 0-initialized regressor is proposed to stabilize fine-tuning in RUL prediction.
\end{highlights}


\maketitle

\section{Introduction}\label{section: Introduction}
Remaining Useful Life (RUL) prediction aims to quantitatively estimate the remaining effective time of equipment, providing objective basis for maintenance decisions or replacement plans\cite{rul_overall_1}. As a key means of avoiding maintenance surplus and deficiency, RUL prediction constitutes the core function of the new generation of Prognostics and Health Management (PHM)\cite{rul_overall_3}. In recent years, deep-learning based RUL prediction methods have attracted increasing attention for their ability to establish accurate end-to-end mapping between monitoring signals and the RUL label without reliance on any expert knowledge\cite{PHM_overall, rul_overall_2}. What these methods primarily rely on is a substantial amount of full-lifecycle degradation data for sufficient model training. However, collecting such run-to-failure full-lifecycle data in real industrial settings is always costly and challenging, which significantly hinders the practical adoption of deep-learning based RUL prediction methods. 

To solve the RUL prediction problem under few-shot domains, transfer learning, represented by domain adaptation\cite{domain_adaption_overall_1, domain_adaption_Cui, domain_adaption_Ding, domain_adaption_Li, domain_adaption_Ragab, domain_adaption_Shang, domain_adaption_Wu} and meta-learning\cite{meta-learning_Schimitt, meta-learning_Cao, meta-learning_Chang, meta-learning_She, meta-learning_Ding, meta-learning_Zhuang}, has become a mainstream approach. The core idea of transfer learning is to leverage the prior knowledge of source domain to target domain by constructing an effective transfer pathway. Although extensive transfer learning-based methods have been developed for few-shot RUL prediction, the majority are limited to transfer across different operating conditions of similar equipment. In other words, the focus of existing methods is on identifying consistent degradation pattern of similar devices under distinct operation conditions. Thus, a prerequisite of these methods is sufficient degradation data from similar equipment can be available to construct the source domain for transfer, which is still a stringent requirement in many practical applications. When historical data of the target equipment is scarce, it may be also difficult to collect a substantial amount of data from similar equipment. For example, the operational lifespan of many types of industrial assets extends for decades, making full lifecycle data collection a long and costly process. Another example is the newly developed or small batch customized equipment domain. The historical degradation data of such types of equipment are considerably limited due to the short service time and small quantity. To sum up, existing few-shot RUL prediction methods based on the idea of device-specific degradation pattern transfer will become infeasible when sufficient degradation data with similar degradation pattern from homogeneous equipment cannot be obtained, which significantly undermines their values in practical application. 

As we know, degradation data for RUL prediction are predominantly time series data, thereby recent advances in time-series representation learning offer a promising way to overcome above limitations. Time series representation models trained in a self-supervised manner can learn generalizable temporal feature representations from large-scale time series datasets. It has been proved that the pre-trained time series representation models have strong generalization ability in cross-domain downstream classification and prediction tasks\cite{SimMTM, TF-C, FEI}. Consequently, leveraging the excellent generalization ability of time series representation model to RUL prediction has the potential to significantly improve the accuracy and robustness of prediction result, especially in few-shot scenes. Moreover, the pre-training process of representation model is not tied to a specific downstream task. This decouples the transfer process from the requirement for equipment-specific degradation data in existing few-shot RUL prediction methods, enabling the transfer of knowledge between domains across diverse and heterogeneous time series.

Nevertheless, time-series representation models for few-shot RUL prediction faces the following critical challenges:

\begin{itemize}
    \item How to leverage widely used univariate time-series representation model to multivariate RUL prediction task? Currently, cross-domain time-series representation learning methods are often designed to capture temporal or spectral features of a single variable\cite{SimMTM, FEI, TS2Vec}, where downstream cross-variable dependencies are not defined and are difficult to define during pre-training. However, in practical applications, complex equipment status is typically characterized by multiple correlated sensor variables. This mismatch between pre-training and downstream task requirements gives rise to the first challenge that must be addressed.
    \item How to design an effective fine-tuning strategy that guarantees stable convergence with extremely limited samples from target equipment for RUL prediction task? Current time-series representation models are parameter-intensive. In few-shot RUL prediction, naively fine-tuning all parameters not only leads to severe overfitting but may also substantially degrade the generalizable priors acquired during pre-training. The application of parameter-efficient finetuning approaches to multivariate RUL prediction has never been explored, which constitutes the second key challenge addressed.
\end{itemize}

Building on above discussion, this paper proposes PEFT-MuTS, a novel Parameter-Efficient Fine-Tuning framework for few-shot Multivariate Time-Series RUL prediction. PEFT-MuTS is built on a general time series representation model, which can be pre-trained on large-scale cross-domain time series data. Then, the pre-trained model is fine-tuned to adopt to the downstream RUL prediction task through an Independent Feature Tuning Network and Meta-Variable-based Low Rank Feature Fusion mechanism with minimal parameters. The main contributions of this paper are as follows:
\begin{itemize}
    \item This paper demonstrates that cross-domain time-series representation models are highly effective for RUL prediction under extremely limited degradation data. By shifting the research focus from “identifying similar degradation patterns” to “leveraging generalizable temporal priors,” the proposed paradigm substantially broadens the scope of transferable source-domain data while simultaneously improving prediction accuracy, thereby offering a new perspective on the few-shot RUL prediction problem. 
    \item A novel parameter-efficient fine-tuning framework, named PEFT-MuTS, is proposed for multivariate few-shot RUL prediction. PEFT-MuTS consists of an Independent Feature Tuning Network and a Meta-Variable–based Low-Rank Feature Fusion mechanism. The former allows the univariate backbone to process each data variable independently, ensuring that representation capacity is not compromised by multivariate interactions. The latter introduces additional learnable meta-variable and employs a lightweight gated network with a small number of parameters to dynamically extract multivariate information, thereby achieving effective adaptation of univariate backbone networks to downstream multivariate RUL prediction tasks.
    \item The unstable phenomena of the fine-tuning process is analyzed in the case of extremely few samples, and a zero-initialized regressor is proposed as a simple yet effective solution to improve the fine-tuning stability, which provides a solid foundation for further researches based on pre-trained representation models for few-shot RUL prediction.
\end{itemize}

\section{Related Works}
\subsection{Few-shot RUL Prediction Methods}
In recent years, although data-driven RUL prediction methods have achieved high accuracy, their reliance on data quality and quantity remains a key challenge. With the rise of transfer learning, few-shot RUL prediction---particularly domain adaptation and meta-learning---has gained increasing attention.

Domain adaptation leverages labeled in-domain data to align source and target feature distributions through loss function constraints, enabling generalization to unlabeled target data. Early approaches used adversarial training to address distribution shifts\cite{domain_adaption_Ragab}. Wu et al.\cite{domain_adaption_Wu} further incorporated limited target labels to enhance adaptation. Ding et al.\cite{domain_adaption_Ding} adopted similar strategies for rotating machinery under condition-induced shifts. Cui et al.\cite{domain_adaption_Cui} employed digital twins to generate full-lifecycle synthetic source data, enabling adaptation to real-world few-shot targets. Li et al.\cite{domain_adaption_Li} advanced this by removing source labels entirely, proposing a self-supervised framework. To address misalignment, Shang et al.\cite{domain_adaption_Shang} introduced a source distillation weighting mechanism that prioritizes source samples closer to the target domain. Mao et al.\cite{fewshotNew2} proposed a wavelet scattering adaptation framework that exploits stage-specific frequency information and domain-adversarial learning to improve cross-machine few-shot RUL prediction while providing uncertainty estimation.

Meta-learning aims to train models that generalize across tasks using few-shot data from multiple domains. A widely used approach is Model-Agnostic Meta-Learning (MAML), which learns a global parameter initialization that enables fast adaptation via support sets. Schmitt et al.\cite{meta-learning_Schimitt} and Ding et al.\cite{meta-learning_Ding} applied MAML to battery and variable-condition RUL prediction. Chang et al.\cite{meta-learning_Chang} incorporated Bayesian modeling into MAML for uncertainty estimation. Cao et al.\cite{meta-learning_Cao} applied MAML to wind turbine bearings by building support sets under varying conditions, while Zhuang et al.\cite{meta-learning_Zhuang} used semantic attention with diverse-condition samples for cross-condition prediction.

Domain adaptation generally assumes the source and target domains share broad distribution similarities--e.g., data from similar equipments--limiting its applicability when source data are scarce. Likewise, MAML-based methods require meta-training tasks constructed from samples that cover sufficiently diverse degradation scenarios, and therefore still rely on a relatively large amount of data from similar equipment. As a result, as discussed in the Section \ref{section: Introduction}, existing methods primarily focus on identifying transferable degradation patterns of similar equipment under different operating conditions. When the available data from similar equipment are insufficient, the transfer process of such methods becomes inherently constrained. Recent advances in time series representation learning offer a promising way to address this limitation.

Overall, existing few-shot RUL prediction methods are generally degradation-oriented, where transferable knowledge is learned from degradation representations with the goal of identifying degradation patterns similar to those of the target equipment. As a result, their effectiveness largely depends on the availability of sufficient degradation data from related equipment or operating conditions. When such similar degradation data are limited, the transfer capability of these methods is inevitably restricted.

In contrast, the proposed framework adopts a fundamentally different transfer paradigm. Instead of modeling degradation-specific characteristics, it builds upon a general-purpose time-series representation model that captures universal temporal patterns shared across diverse time-series data. Consequently, the transferable knowledge is no longer confined to degradation trajectories, but can be acquired from abundant cross-domain time-series datasets that are easier to obtain. This broader transfer pathway significantly alleviates the dependence on domain-specific degradation data, thereby reducing the difficulty of developing RUL prediction models under few-shot conditions.

\subsection{Time Series Representation Learning}

Representation learning is a cornerstone of deep learning, enabling models to acquire task-agnostic high-level semantic features through self-supervised pre-training on large datasets, which are then transferred to downstream tasks to reduce data demand and improve convergence\cite{TF-C}\cite{FEI}. Due to the high variability of time series data, it is challenging to construct time series representation models with strong generalizability, and related research started relatively late. Notable approaches such as TF-C\cite{TF-C} aim to extract consistent and discriminative features from both the time and frequency domains, leveraging diverse augmentation strategies to enhance transferability across domains. TS2Vec\cite{TS2Vec} employs contrastive learning to generate positive and negative pairs from time series data, learning discriminative representations with strong transfer performance. SimMTM\cite{SimMTM}, grounded in manifold theory, uses masked modeling and learns consistent and diverse representations across two feature subspaces. TimeMAE\cite{fundationModelNew1} further reformulates masked time-series modeling by treating short sub-series as semantic units rather than individual timestamps. It adopts a decoupled masked autoencoder together with complementary reconstruction objectives, leading to more transferable representations under label-scarce and cross-domain settings. FEI\cite{FEI} corrects inappropriate priors in contrastive-based time series representation models and builds a framework better suited for learning semantic continuity, thus improving generalization.

\subsection{Parameter-efficient Fine-tuning}
\label{sec: PEFT work and our goal}
Although strong representation models enhance downstream transferability, preventing overfitting in low-data regimes becomes imperative, especially when adapting large pre-trained models. PEFT aims to improve fine-tuning performance with minimal additional parameters, and has become foundational in CV and NLP tasks\cite{PEFT}. Techniques like Side Tuning\cite{Side_Tuning_1, Side_Tuning_2, DTL}, P-Tuning\cite{p-tuning1, p-tuning2, p-tuning3} and LoRA\cite{LoRA_1, LoRA_2} are representative. Recent studies have also begun investigating fine-tuning strategies for time-series foundation models. For example, MSFT\cite{peftNew1} explicitly incorporates multi-scale temporal modeling into the fine-tuning process of encoder-based time-series foundation models, demonstrating that carefully designed fine-tuning strategies can substantially outperform both naive full fine-tuning and conventional PEFT methods. However, most existing PEFT methods are developed primarily for the CV and NLP domains and are tightly coupled with standard Transformer architectures. For example, the widely used LoRA\cite{LoRA_1}\cite{LoRA_2} fine-tuning strategy is almost exclusively tied to the MLP layers of Transformer architectures, while P-Tuning\cite{p-tuning1}\cite{p-tuning2} and Prompt Tuning\cite{p-tuning3} is bound to the self-attention mechanism. In contrast, models for time-series representation learning and RUL prediction are not restricted to the standard Transformer paradigm and face a fundamental challenge arising from discrepancies in data variables between pre-training and downstream tasks\cite{FEI}. Consequently, PEFT architectures tailored for multivariate RUL prediction based on time-series representation models remain underexplored and warrant further investigation.

\section{Proposed Method}
\subsection{Formulation of the problem}
Given a set of multivariate operational monitoring data $X \in \mathbb{R}^{B \times N \times T}$, where $B$ is the number of available data samples, $N$ is the number of monitored variables per device, and $T$ is the time length of each sample, the corresponding device lifespans are denoted by $Y \in \mathbb{R}^B$. Here, the RUL label $y_i$ is defined at the final timestamp $T_t$ of each sample $x_i$.

Assuming the ideal full life-cycle length of a device is $L$, this study focuses on the scenario where $B \ll L$, indicating that the available monitoring data cannot fully cover the complete degradation process of the device.

In the standard data-driven RUL prediction framework, the training objective is typically defined as\cite{Dual-Mixer}\cite{IMDSSN}:
\begin{equation}
    \mathcal{L}(\theta, \phi) = \frac{1}{2B} \left\| Y - f_{\theta, \phi}(X) \right\|_2^2
    \label{eq:general_loss}
\end{equation}
where:
\begin{equation}
    f_{\theta, \phi}(X) = h_\phi(g_\theta(X))
    \label{eq:genral compute form}
\end{equation}

Here, $g_\theta$ denotes a randomly initialized feature extractor, and $h_\phi$ is a randomly initialized RUL regression head. Based on this setup, the proposed PEFT-MuTS architecture introduces a pre-trained backbone, a fine-tuning scheme, and a zero-initialized regression head for enhanced performance under limited degradation data.

\subsection{PEFT-MuTS Framework}
\begin{figure}
    \centering
    \includegraphics[width=0.65\linewidth]{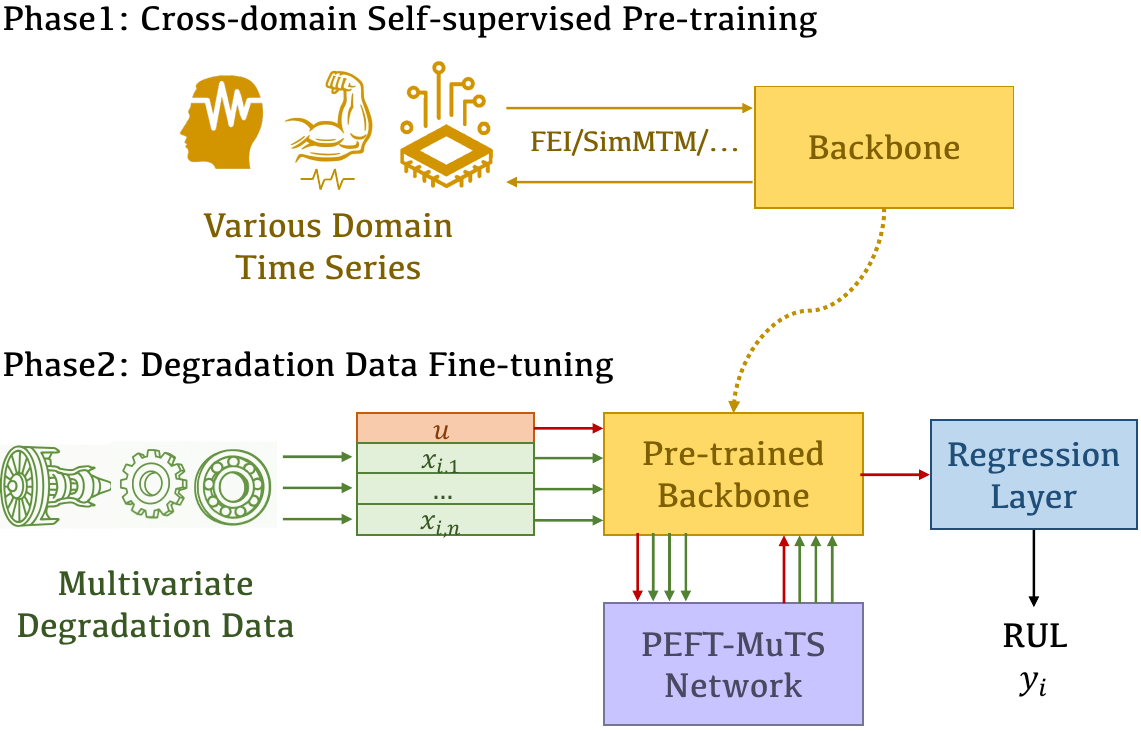}
    \caption{The Overall framework of the proposed PEFT-MuTS.}
    \label{fig:overall}
\end{figure}

The overall architecture of PEFT-MuTS is illustrated in Fig.~\ref{fig:overall}. PEFT-MuTS operates in two stages:

\begin{itemize}
    \item Stage 1 conducts self-supervised pre-training of the backbone network $g_\theta$ using abundant and readily available cross-domain univariate time series data, leveraging existing methods such as FEI and SimMTM.
    
    \item Stage 2 performs fine-tuning, where the backbone network is adapted to the downstream RUL prediction task.
\end{itemize}

During the pre-training stage (Stage 1), unlike conventional transfer learning for RUL prediction, the objective is not to learn degradation-specific knowledge, but to acquire general-purpose time-series representations through self-supervised learning. Such pre-training captures universal temporal characteristics, including trends, periodicity, frequency patterns, and local temporal dependencies, which are widely shared across different time-series domains. Since equipment degradation is ultimately manifested through the evolution of these temporal characteristics, the learned representations can be effectively transferred to downstream degradation modeling, even when the pre-training data are unrelated to equipment health. 


During the fine-tuning stage (Stage 2), the parameters of the pre-trained backbone network are frozen, and the PEFT-MuTS network is introduced as the tunable component for the backbone. Only the parameters of PEFT-MuTS are updated through gradient descent.

Specifically, a PEFT-MuTS network layer consists of an Independent Feature Tuning Network and a Meta-Variable–based Low-Rank Feature Fusion module, as illustrated in Fig.~\ref{fig:layer details}. In the figure, the green “Variable Path” denotes the processing path of the Independent Feature Tuning Network, which handles each data variable independently. The red “Meta-Variable Path” corresponds to the Low-Rank Feature Fusion, where information from different variables is aggregated.

The PEFT-MuTS layer performs feature extraction in parallel with the pre-trained backbone network, taking the same input as the backbone and feeding its activate output back into the backbone. The final output representation relies solely on the meta-variable features, which progressively fuse information from other data variables across layers. In addition, a “zero-initialized” regressor is introduced to stabilize fine-tuning under few-shot settings.

The computational details of each component are presented next. Section \ref{sec:Independent Feature Tuning Network} introduces the Independent Feature Tuning Network, Section \ref{sec:Meta-Variable-based Low Rank Feature Fusion} describes the Meta-Variable–based Low-Rank Feature Fusion, and Section \ref{section: zero-initialized} analyzes the fine-tuning stability issue and addresses it through the proposed simple yet effective zero-initialized regressor.


\subsubsection{Independent Feature Tuning Network}\label{sec:Independent Feature Tuning Network}
\begin{figure}
    \centering
    \includegraphics[width=0.75\linewidth]{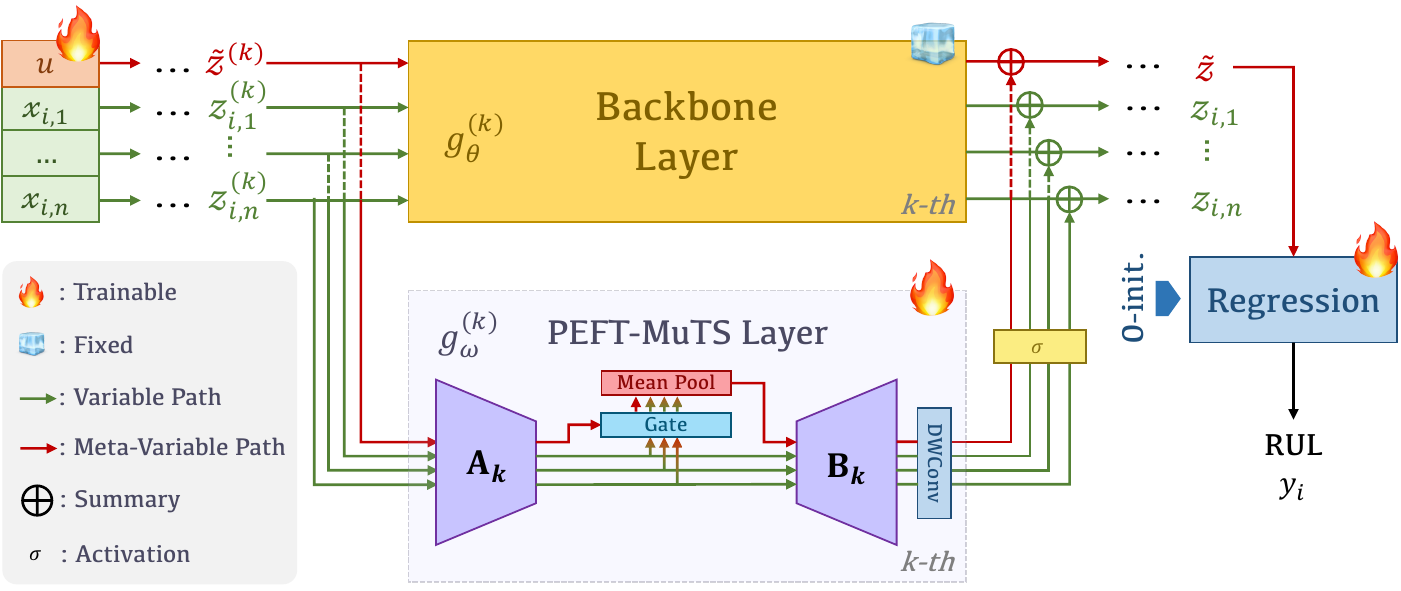}
    \caption{The details of  PEFT-MuTS framework.}
    \label{fig:layer details}
\end{figure}
The design of the Independent Feature Tuning Network is motivated by the need to preserve the univariate nature of the pre-trained time-series backbone. Since the backbone is pre-trained on univariate data, its prior structure is inherently tailored to feature extraction from a single time series. Naively introducing multivariate information would disrupt this pre-trained prior, potentially rendering the extracted features unreliable. Therefore, to preserve the backbone’s prior structure while keeping the number of trainable parameters minimal during fine-tuning, an Independent Feature Tuning Network is introduced. This module enforces variable-wise feature extraction and employs low-rank mappings inspired by LoRA\cite{LoRA_1}. This module serves as the fundamental component for addressing challenge 1 discussed in the Section \ref{section: Introduction}.

As illustrated in Fig. \ref{fig:layer details}, given a multivariate degradation sample $x_i \in \mathbb{R}^{N \times T}$ with $N$ variables, the computation at the k-th layer with the tuning module is as follows:
\begin{equation}
    z_i^{(k+1)} = g_{\theta}^{(k)}(z_i^{(k)}) + \text{SiLU}(g_{\omega}^{(k)}(z_i^{(k)}))
    \label{eq:xi process overall}
\end{equation}
where $z_i^{(k)} \in \mathbb{R}^{N \times d_k}$ is the intermediate feature at the $k$-th layer with dimension $d_k$, and $z_i^{(1)} = x_i$. SiLU represents the Sigmoid Linear Unit activation function\cite{silu}, is used to regulate the overall information flow to the backbone layer and is defined as:
\begin{equation}
    \text{SiLU}(x) = x * \text{Sigmoid}(x) = x * \frac{1}{1 + e^{-x}}
\end{equation}
$g_{\theta}^{(k)}$ denotes the $k$-th layer of the backbone, and $g_{\omega}^{(k)}$ is the corresponding tuning layer, defined as:

\begin{equation}
    g_{\omega}^{(k)}(z_i^{(k)}) = \mathrm{Conv}\left(  z_i^{(k)} \mathbf{A}_k \mathbf{B}_k \right)
    \label{eq:xi process detail}
\end{equation}
Here, $\mathbf{A}_k \in \mathbb{R}^{d_k \times r_k}$ and $\mathbf{B}_k \in \mathbb{R}^{r_k \times d_k}$ are the core parameters of the tuning module, where $r_k$ denotes the projected dimension and is always much smaller than $d_k$, i.e., $r_k \ll d_k$. This low-rank mapping design ensures a minimal parameter overhead even under high-dimensional feature tuning. $\mathbf{A}_k$ is randomly initialized, while $\mathbf{B}_k$ is zero-initialized, encouraging the tuning output to start at zero and enabling residual learning that preserves pre-trained features and reduces overfitting.

\subsubsection{Meta-Variable-based Low Rank Feature Fusion}\label{sec:Meta-Variable-based Low Rank Feature Fusion}
As mentioned in the previous subsection, the Independent Feature Tuning Network ensures that the feature information of each variable is preserved during inter-layer propagation. However, how to achieve multivariate fusion on this basis is the focus of this subsection. To this end, we propose a Meta-Variable–based Low-Rank Feature Fusion mechanism to further address Challenge 1 discussed in the Section \ref{section: Introduction}, building upon the Independent Feature Tuning Network.

In this mechanism, we introduce an independently learned variable, referred to as a meta-variable, denoted by $u \in \mathbb{R}^{1 \times T}$. The meta-variable is specifically designed as a dedicated variable solely for information fusion. As part of the tunable parameters, it can be updated according to the requirements of the task and the backbone network. Within the low-rank space of PEFT-MuTS, this variable dynamically receives feature information from the data variables. This design preserves the independence of the data variables during their processing while ensuring that the features of $u$ remain controllable during inter-layer propagation through its adjustable characteristics. Moreover, performing the fusion within a low-rank space keeps the number of fusion-related parameters manageable and promotes a more purposeful low-rank mapping $\mathbf{A}_k$, thereby constructing a subspace during optimization that is better suited for downstream tasks.

Denote the meta-variable as $u \in \mathbb{R}^{1 \times T}$, as illustrated by the red line in Figure \ref{fig:layer details}. It is initialized as a zero vector with the same sequence length as the input $x_i$, enabling parallel processing with the input sequence: $\{u, x_i\} \in \mathbb{R}^{(N+1) \times T}$. 
Let the low-rank representation of the intermediate features of the data variables and the meta-variable be denoted by
\begin{equation}
v^{(k)}_i = \{\tilde{z}^{(k)}, z^{(k)}_i\} \mathbf{A}_k \in \mathbb{R}^{(N+1)\times r_k}
\end{equation}
where $\tilde{z}^{(k)}$ represents the intermediate representation of the meta-variable $u$, with $\tilde{z}^{(1)} = u$. The processing of the meta-variable is then formulated as
\begin{equation}
    \tilde{z}^{(k+1)} = g_\theta(\tilde{z}^{(k)}) + \text{SiLU}(f^{(k)}(v^{(k)}_i) \mathbf{B}_k)
    \label{eq:u process detail}
\end{equation}
where $f^{(k)}$ denotes the fusion function, defined by
\begin{equation}
    f^{(k)}(v^{(k)}_i) = \text{MeanPool}(\text{Gate}(v^{(k)}_i))
\end{equation}
Here, $\text{MeanPool}$ represents an average pooling operation along the variable dimension with a window size of N+1, aggregating the filtered features of all data variables and the meta-variable. $\text{Gate}$ is a self-gating mechanism that filters the intermediate features of all data variables and the meta-variable, selecting the parts used for multivariate fusion:
\begin{equation}
    \text{Gate}(v^{(k)}_i) = \text{Sigmoid}(v^{(k)}_i \mathbf{W}_k) * v^{(k)}_i
\end{equation}
where $W_k \in \mathbb{R}^{r_k \times r_k}$ is a learnable parameter updated along with the model.

It should be noted that the above operations are applied exclusively to the intermediate variables of the meta-variable $u$. The resulting $\tilde{z}^{(k+1)}$ still only represents the updated meta-variable, while the data variables continue to follow the processing pipeline defined in Equation \ref{eq:xi process detail}.

Overall, the meta-variable $u$ participates in the network in the same manner as the remaining data variables and is processed identically by the backbone. The only differences lie in its learnable nature and its dedicated role during variable fusion. Together, these properties decouple the fusion process from the data variables, while preserving the original feature representations extracted by the backbone for each variable.

In this way, feature fusion of the data variables is achieved through the meta-variable $u$ while maintaining the independent processing of each data variable. The gating mechanism constructs a flexible fusion path. The next subsection will analyze issues related to fine-tuning stability.

\subsection{Zero-initialized RUL Regressor}
\label{section: zero-initialized}

In practice, we observe that fine-tuning pre-trained time series models on extremely small samples often results in highly unstable outcomes. Repeated training runs may yield significantly different results, and in some cases, the model may become trapped in poor local minima where the loss hardly decreases. To better understand this phenomenon, we conduct a variance-based analysis. 

Consider the RUL prediction model trained with batch gradient descent as in \eqref{eq:genral compute form}. Given a linear regressor $h_\phi$, where $\phi \sim \mathcal{N}(0, \sigma_\phi^2)$, and extracted degradation features $z \in \mathbb{R}^{B \times d}$ with $z \sim \mathcal{N}(0, \sigma_z^2)$, the prediction is $\hat{Y} = z \phi$, while the ground truth $Y \in \mathbb{R}^{B}$ is assumed to follow $Y \sim \mathcal{N}(0, \sigma_Y^2)$. Using the $L_2$ loss, the gradient update of the $j$-th parameter $\phi_j$ is:
\begin{equation}
\Delta \phi_j = -\eta \frac{\partial \mathcal{L}}{\partial \phi_j} = -\frac{\eta}{B} \sum_{i=1}^{B} z_{ij}(Y_i - z_i \phi)
\end{equation}

Let $E_i = Y_i - z_i\phi$ be the prediction error. Since $\phi$ is randomly initialized, $E_i$ is also a random variable. Its variance can be derived as:
\begin{equation}
    \text{Var}(E_i) = \text{Var}(Y_i - z_i \phi)= \sigma_Y^2 + d \cdot \sigma_z^2 \cdot \sigma_\phi^2
\end{equation}

Then, the variance of the parameter update $\Delta \phi_j$ becomes:
\begin{equation}
    \text{Var}(\Delta \phi_j) = \text{Var} \left( -\frac{\eta}{B} \sum_{i=1}^{B} z_{ij} E_i \right) = \frac{\eta^2}{B^2} \sum_{i=1}^{B} \text{Var}(z_{ij} E_i)
\end{equation}

For simplicity, assuming independence between $z_{ij}$ and $E_i$, we have:
\begin{equation}
    \text{Var}(z_{ij} E_i) = \text{Var}(z_{ij}) \cdot \text{Var}(E_i) = \sigma_z^2 \cdot (\sigma_Y^2 + d \cdot \sigma_z^2 \cdot \sigma_\phi^2)
\end{equation}

Thus:
\begin{equation}
    \text{Var}(\Delta \phi_j) = \frac{\eta^2}{B} \cdot \sigma_z^2 \cdot (\sigma_Y^2 + d \cdot \sigma_z^2 \cdot \sigma_\phi^2)
    \label{eq:unstable analysis}
\end{equation}

This analysis shows that the gradient variance is positively correlated with the variance of the input features $\sigma_z^2$, the label variance $\sigma_Y^2$, and the initialization variance of the regressor $\sigma_\phi^2$. A large update variance leads to strong randomness in the early training stage, manifesting as instability across different training runs.

Notably, since pre-trained representation models produce more discriminative features in the early stage of fine-tuning, the initial feature variance $\sigma_z^2$ is usually high. This leads to larger $\text{Var}(\Delta \phi)$ and hence greater instability compared to training from scratch.

To suppress this effect, we adopt zero-initialized regressor in PEFT-MuTS, i.e., $\phi = \mathbf{0}$, which yields the minimal initial gradient variance:
\begin{equation}
    \text{Var}(\Delta \phi_j) = \frac{\eta^2}{B^2} \cdot \sigma_z^2 \cdot \sigma_Y^2
\end{equation}

This eliminates the contribution of $\sigma_\phi^2$, thereby avoiding additional instability caused by common initializers such as Xavier\cite{xavier} or Kaiming\cite{kaiming}, and significantly reduces the impact of $\sigma_z^2$. 

It should be noted that the above derivation is intended to provide an intuitive analysis of the optimization dynamics rather than a rigorous theoretical proof. For analytical tractability, several simplifying assumptions are adopted, including Gaussian feature distributions and independence between the extracted features and the prediction error. Although these assumptions may not strictly hold throughout the entire training process, they enable the dominant factors affecting the gradient variance during the early stage of fine-tuning to be explicitly characterized. Moreover, the analysis is specifically derived for fine-tuning with a linear prediction head and may not directly extend to architectures employing nonlinear prediction heads. Accordingly, the analysis primarily explains the optimization behavior during the early stage of fine-tuning rather than the complete optimization dynamics.

Based on this analysis, a zero-initialized regressor is introduced to address this issue, thereby resolving Challenge 2 discussed in Section \ref{section: Introduction}.In subsequent experiments, we empirically validate the effectiveness of zero-initialized regressor in improving fine-tuning stability.

\section{Experiments}
\subsection{Preparing Datasets}
\subsubsection{Basic information}
This study employs two widely-used prognostic datasets: the C-MAPSS dataset\cite{cmapss} for aircraft engine degradation and the XJTU-SY Bearing dataset\cite{bearing} for bearing degradation.  

For the C-MAPSS dataset, we adopt the FD002 and FD004 subsets, which each contain six operating conditions and represent more realistic and complex operating scenarios. We select 14 commonly used sensor variables that exhibit degradation trends\cite{Dual-Mixer}, as shown in Table~\ref{tab:features}. More details about the C-MAPSS dataset can be found in reference \cite{cmapss}.  

For the Bearing dataset, we use all three operating condition subsets, denoted as OP-A, OP-B, and OP-C, which correspond to different speeds and loads. Each sample contains vibration signals collected from both horizontal and vertical directions, as shown in Table~\ref{tab:features}. More details about the Bearing dataset can be found in \cite{bearing}.

\begin{table*}[t]
	\caption{Selected input features for different datasets.} 
    \label{tab:features}
	\centering
	\begin{tabular}{c|c|l}
		\toprule
		\textbf{Dataset} & \textbf{Symbol} & \textbf{Description}\\
		\midrule
        \multirow{14}{*}{C-MAPSS}
		&T24	&Total temperature at LPC outlet\\
		&T30	&Total temperature at HPC outlet\\
		&T50	&Total temperature at LPT outlet\\
		&P30	&Total pressure at HPC outlet\\
		&Nf	&Physical fan speed\\
		&Nc	&Physical core speed\\
		&Ps30	&Static pressure at HPC outlet\\
		&Phi	&Ratio of fuel flow to Ps30\\
		&NRf	&Corrected fan speed\\
		&NRc	&Corrected core speed\\
		&BPR	&Bypass ratio\\
		&htBleed	&Bleed enthalpy\\
		&W31	&HPT coolant bleed\\
		&W32	&LPT coolant bleed\\
		\midrule
        \multirow{2}{*}{XJTU-SY Bearing}
		&$X_h$	&Horizontal vibration acceleration signal\\
		&$X_v$	&Vertical vibration acceleration signal\\
		\bottomrule
	\end{tabular}
\end{table*}

\subsubsection{RUL Labels}  
For the C-MAPSS dataset, we adopt the commonly used piecewise linear labeling method\cite{Dual-Mixer}. The degradation start point is defined as 120 cycles before failure. Prior to this point, the unit is considered healthy, and RUL degrades linearly afterward.  
For the Bearing dataset, we follow Su et al.~\cite{bearing_fault_point_ref_1}\cite{meta-learning_Chang} and use the root mean square (RMS) method combined with the three-sigma rule to determine the degradation onset. Similarly, a linear degradation is assumed after the degradation point.  
All RUL labels are normalized to the range [0, 1], where 1 denotes the healthy state and 0 denotes failure\cite{meta-learning_Chang}\cite{Dual-Mixer}.

\subsubsection{Construction of Few-shot Datasets}  
Based on the above datasets, we construct small-sample degradation datasets of varying scales using the following three-step procedure:

First, a sliding window of fixed length is applied to generate time series segments, with the RUL assigned as the value at the last time step. This creates sample-label pairs across the full degradation cycle. For C-MAPSS, the window size is 30 with a step of 15; for the Bearing dataset, the window size is 1024 with a step of 32768.

Secondly, since the original C-MAPSS and Bearing datasets consist of multiple devices with complete life-cycle data, we construct the few-shot datasets through a three-step random sampling procedure: Device Retention, RUL Coverage, and Global Data Retention. These three steps progressively reduce the amount of available training data from the device level, life-cycle level, and sample level, respectively. Specifically, Device Retention selects the device units to be included in the few-shot dataset. RUL Coverage then determines how much of the life-cycle data from the selected devices is preserved to generate fragmented life-cycle trajectories. Finally, Global Data Retention further randomly samples from the remaining data to obtain the final few-shot dataset. The detailed definitions are as follows:

\begin{enumerate}
\item \textbf{Device Retention Percentage $p_1$}: For datasets with multiple device units (e.g., C-MAPSS), entire device units are retained with probability $p_1$, while samples from the remaining devices are discarded. This step is applied only to C-MAPSS. For the Bearing dataset, which contains only five units under each operating condition (OP-A/B/C), we directly use the first unit from each condition.
\item \textbf{RUL Coverage Percentage $p_2$}: A subset of RUL values is randomly selected with probability $p_2$ (without replacement), and all samples corresponding to the selected RUL values are retained. Healthy samples (RUL = 1) are always preserved due to their abundance and accessibility in practical scenarios.
\item \textbf{Global Data Retention Percentage $p_3$}: To introduce additional randomness and better simulate practical conditions, the remaining samples are further retained with probability $p_3$. A smaller value of $p_3$ results in fewer training samples.
\end{enumerate}

Table~\ref{tab:datasets} summarizes the resulting small-sample training sets. Since the OP-A operating condition in the Bearing dataset contains relatively few samples, we divide this condition into only two few-shot scenarios. In the Data Proportion columns, the Device, RUL, and Global Retention values indicate the respective percentages, while the Total column shows the overall percentage of each subset relative to the original dataset. In the Number of Samples columns, Early Stage, Middle Stage, and Late Stage report the number of samples with RUL < 1, divided into RUL$\in[0.7, 1.0)$, RUL$\in[0.3, 0.7)$, and RUL$\in[0.0, 0.3)$ respectively. The Health Stage column indicates the number of samples with RUL = 1.
\begin{table}
\centering
\caption{Dataset statistics with varying proportions. }
\begin{tabular}{l|cccc|cccc}
\toprule
\textbf{Dataset} 
& \multicolumn{4}{c|}{\textbf{Data Percentage}} 
& \multicolumn{4}{c}{\textbf{Number of Training Samples}}
\\
& Device $p_1$ & RUL $p_2$ & Global $p_3$ & Total & Early & Middle & Late & Health\\
\midrule
FD002-1  & 30\% & 3\%  & 80\% & 0.72\%  & 0 & 8 & 5   & 270 \\
FD002-2  & 30\% & 8\%  & 80\% & 1.92\%  & 0 & 11 & 28 & 220 \\
FD002-3  & 30\% & 30\% & 80\% & 7.20\%  & 37 & 54 & 44 & 243 \\
FD002-4  & 30\% & 60\% & 80\% & 14.40\% & 100 & 114 & 76 & 248 \\
\midrule
FD004-1  & 30\% & 3\%  & 80\% & 0.72\%  & 0 & 2 & 8   & 371 \\
FD004-2  & 30\% & 8\%  & 80\% & 1.92\%  & 6 & 12 & 25 & 337 \\
FD004-3  & 30\% & 30\% & 80\% & 7.20\%  & 54 & 40 & 34 & 335 \\
FD004-4  & 30\% & 60\% & 80\% & 14.40\% & 67 & 129 & 95 & 327 \\
\midrule
OP-A-1   & /    & 5\%  & 100\% & 5\%  & 0 & 0 & 1 & 77  \\
OP-A-2   & /    & 10\% & 100\% & 10\% & 0 & 1 & 2 & 77  \\
\midrule
OP-B-1   & /    & 10\% & 10\%  & 1\%  & 0 & 0 & 1 & 44  \\
OP-B-2   & /    & 10\% & 50\%  & 5\%  & 1 & 0 & 1 & 226 \\
OP-B-3   & /    & 10\% & 100\% & 10\% & 0 & 1 & 1 & 454 \\
\midrule
OP-C-1   & /    & 10\% & 10\%  & 1\%  & 0 & 0 & 1 & 33  \\
OP-C-2   & /    & 10\% & 50\%  & 5\%  & 0 & 1 & 1 & 169 \\
OP-C-3   & /    & 10\% & 100\% & 10\% & 0 & 1 & 1 & 340 \\
\bottomrule
\end{tabular}
\label{tab:datasets}
\end{table}

\subsubsection{Test Sets}  
The datasets in Table~\ref{tab:datasets} are used for training.  
For testing, we use the original test sets.  
FD002 contains 259 test engine instances, and FD004 has 248 test engines, all with incomplete degradation.  
For the Bearing dataset, we use all remaining bearings (units 2-5) from OP-A, OP-B, and OP-C as the test set.  
No additional processing is performed on test sets, and window sizes and step lengths are consistent with the training set.

\subsubsection{Normalization}  
Following standard practice, we apply min-max normalization to all datasets.  
Each independently constructed small-sample training set is normalized separately, using its own min and max values. These values are then applied to normalize the corresponding test set to prevent information leakage during training.

\subsection{Implementation Details}


PEFT-MuTS employs a 1-D ResNet-18 backbone that has been pre-trained using the FEI method \cite{FEI}. The pre-training configuration strictly follows the original paper’s settings, and the process is conducted on the SleepEEG dataset\cite{SleepEEG}. The training set of this dataset contains 371,055 EEG signal samples, each with a length of 178, covering a total of 66,047,790 time points, sampled at 100 Hz. SleepEEG is selected because it exhibits diverse temporal patterns spanning multiple frequency ranges and has been widely adopted as a benchmark for self-supervised time-series representation learning \cite{SimMTM,TF-C,FEI}. Such temporal diversity enables the encoder to learn more generalizable time-series representations, providing an effective initialization for downstream RUL prediction.

Each residual block of ResNet-18 (excluding the input embedding layer) is followed by a PEFT-MuTS layer. As each block has 2 convolutional layers (excluding shortcuts), 8 PEFT-MuTS layers are attached in total. Table\ref{tab:layer_information} details each module: \textbf{Layers} denotes the corresponding backbone block, $d_k$ is the input feature dimension of the block, which is also the input dimension to the PEFT-MuTS layer, $r_k$ is its inner projection dimension, and \textbf{DWConv} specifies the number of depthwise separable kernels (kernel size 3) for dimensional alignment when needed. This setup adds only ~74.9K parameters--just 0.50\% of the backbone.

\begin{table}
\centering
\caption{PEFT-MuTS structure details with ResNet-18 backbone.}
\setlength{\tabcolsep}{3pt}
\begin{tabular}{c|c|cc}
\toprule
\textbf{Layers} & $d_k$ & $r_k$ & \textbf{DWConv} \\
\midrule
Input Embedding &   $N$ &   /   &   / \\
Block1 &   64 &   128   &   128 \\
Block2 &   128 &   32   &   / \\
Block3 &   128 &   32   &   256 \\
Block4 &   256 &   4   &   / \\
Block5 &   256 &   4   &   512 \\
Block6 &   512 &   2   &   / \\
Block7 &   512 &   2   &   1024 \\
Block8 &   1024 &   1   &   / \\
\midrule
Total Params   &   15M &   \multicolumn{2}{c}{74.9K(0.50\%)} \\
\bottomrule
\end{tabular}
\label{tab:layer_information}
\end{table}
Fine-tuning is conducted for 300 epochs. Due to the extremely small sample size, a validation set is not partitioned, and thus early stopping based on validation loss is not employed. The optimizer used is AdamW with $\beta = (0.9, 0.999)$, an initial learning rate of 0.001, an exponential decay rate of 0.99, and a maximum batch size of 32.

All experiments were conducted on a Dell Precision 7920 Tower workstation equipped with dual Intel Xeon Silver 4210R processors, 64\,GB DDR4 ECC memory, and two NVIDIA RTX 3090 GPUs. GPU~0 was used for all experiments. The software environment consisted of Ubuntu Linux (kernel 5.4.0-150-generic), NVIDIA driver 530.30.02, and CUDA 12.1.

\subsection{Experimental Results}

\subsubsection{Baselines}
We select several supervised learning methods for RUL prediction as baselines, including Dual-Mixer\cite{Dual-Mixer}, DAMCNN\cite{DAMCNN}, IMDSSN\cite{IMDSSN} and ResNet\cite{ResNet-18}, to compare the proposed method with conventional supervised approaches in terms of predictive performance. In addition, we introduce state-of-art domain adaptation method designed for few-shot scenarios TACDA\cite{TACDA} and a meta-learning-based few-shot RUL prediction method MKDPINN\cite{MKDPINN} to evaluate the performance of traditional few-shot learning paradigms under extremely limited data conditions. Furthermore, an Side-Tuning-based fine-tuning framework DTL\cite{DTL} is incorporated to compare the proposed fine-tuning strategy with existing Side Tuning-based approaches in multivariate RUL prediction scenarios.
On this basis, standard fine-tuning strategies are further included as fundamental baselines, namely full fine-tuning (Full) and linear fine-tuning (Linear), to verify the performance gains obtained from the pre-training stage.

To ensure a fair comparison, all baseline methods were implemented using the hyperparameter settings reported in their original publications, with the only exception being the batch size, which was uniformly set to 32 due to the limited number of training samples. As shown in Table~\ref{tab:baseline_hyper}, h, layers, epoch, and lr denote the hidden dimension, the number of model layers, the number of training epochs, and the learning rate, respectively. The remaining hyperparameters are specific to each baseline method, and readers are referred to the corresponding original publications for further details. All baseline methods and PEFT-MuTS were trained and evaluated using the same training/test split and identical random seeds to ensure a fair comparison.

All reported results are presented as the mean and standard deviation over five independent runs. Both the proposed method and all baseline methods were evaluated using the same five random seeds: {20251, 20252, 20253, 20254, 20255}. 
\begin{table}[t]
\centering
\caption{Hyperparameter configurations of all baseline methods.}
\label{tab:baseline_hyper}
\begin{tabular}{l|p{11.7cm}}
\toprule
\textbf{Method} & \textbf{Hyperparameters} \\
\midrule
DTL &
$r=8$, $M=1$, epoch$=300$, lr$=0.001$, batch size$=32$ \\

MKDPINN &
$h=64$, meta batch size$=5$, lr$=0.001$, inner step$=8$, inner batch size$=32$, outer epoch$=200$, step size$=0.1$, derivatives order$=2$, epoch$=100$ \\

TACDA &
$h=32$, layers$=3$, backbone=LSTM, dropout$=0.5$, batch size$=32$, lr$=5\times10^{-5}$, discriminator lr$=5\times10^{-5}$, $\alpha_{\rm nce}=0.2$ \\

DAMCNN &
$h=32$, epoch$=300$, lr$=0.001$, batch size$=32$ \\

IMDSSN &
$h=64$, layers$=2$, heads$=2$, positional encoding=True, epoch$=300$, lr$=0.001$, batch size$=32$ \\

Dual-Mixer &
$h=32$, layers$=6$, epoch$=300$, lr$=0.001$, batch size$=32$ \\
\bottomrule
\end{tabular}
\end{table}


\subsubsection{Evaluation Metrics}
We employ MAE, RMSE, and MAPE as performance evaluation metrics. For the Bearing dataset, due to the presence of numerous samples with zero RUL in the test set, we replace MAPE with SMAPE to avoid numerical explosion caused by division by zero. The formulations of these metrics are as follows::
\begin{equation}
    \mathrm{MAE} = \frac{1}{n} \sum_{i=1}^{n} |y_i - \hat{y}_i|
\end{equation}

\begin{equation}
    \mathrm{RMSE} = \sqrt{ \frac{1}{n} \sum_{i=1}^{n} (y_i - \hat{y}_i)^2 }
\end{equation}

\begin{equation}
    \mathrm{MAPE} = \frac{100\%}{n} \sum_{i=1}^{n} \left| \frac{y_i - \hat{y}_i}{y_i} \right|
\end{equation}

\begin{equation}
    \mathrm{SMAPE} = \frac{100\%}{n} \sum_{i=1}^{n} \frac{|y_i - \hat{y}_i|}{(|y_i| + |\hat{y}_i|)/2}
\end{equation}

\begin{table*}[h]
\centering
\caption{Comparison of RUL prediction performance on small-sample C-MAPSS datasets across different methods.}
\resizebox{\linewidth}{!}{
\setlength{\tabcolsep}{3pt}
\begin{tabular}{cl|c|c|c|c|c|c|c|c|c}
\toprule
\multicolumn{2}{c|}{Methods} & Metrics & FD002-1 & FD002-2 & FD002-3 & FD002-4 & FD004-1 & FD004-2 & FD004-3 & FD004-4 \\
\midrule
\multirow{18}{*}{\rotatebox[origin=c]{90}{\textbf{Transfer Learning}}} 
    & \multirow{3}{*}{\textbf{PEFT-MuTS}}
        & MAE  &\resultpmbf{0.1529}{0.01}&\resultpmbf{0.1391}{0.00}&\resultpmbf{0.1467}{0.01} &\resultpmbf{0.1573}{0.00}&\resultpmbf{0.1072}{0.00} &\resultpmbf{0.1225}{0.01} &\resultpmbf{0.1222}{0.00} &\resultpmbf{0.1469}{0.01} \\
    & & MAPE(\%) &\resultpmbf{38.382}{2.28} &\resultpmbf{28.280}{0.73} &\resultpmbf{27.050}{0.80} &\resultpmbf{26.410}{0.45} &\resultpmbf{26.582}{0.75} &\resultpmbf{26.828}{0.81} &\resultpmbf{23.302}{0.88}&\resultpmul{22.976}{0.29} \\
    & & RMSE &\resultpmbf{0.2415}{0.01} &\resultpmbf{0.2042}{0.00} &\resultpmul{0.2084}{0.01} &\resultpmbf{0.1922}{0.01} &\resultpmbf{0.2082}{0.00} &\resultpmbf{0.2078}{0.00} &\resultpmbf{0.1907}{0.00} &\resultpmbf{0.2047}{0.00} \\
\cmidrule{2-11}
    & \multirow{3}{*}{Full}
        & MAE & \resultpm{0.1831}{0.01}&\resultpm{0.1821}{0.02} &\resultpm{0.1861}{0.01} &\resultpm{0.1792}{0.01} &\resultpmul{0.1228}{0.00} &\resultpm{0.1434}{0.01} &\resultpm{0.1417}{0.00} &\resultpm{0.1748}{0.00} \\
    & & MAPE(\%) &\resultpm{47.466}{1.01} &\resultpmul{35.688}{3.90} &\resultpm{32.160}{0.77} &\resultpm{28.476}{0.53} &\resultpmul{30.870}{0.65} &\resultpmul{29.418}{1.56} &\resultpm{24.726}{0.74} &\resultpm{26.192}{0.33} \\
    & & RMSE &\resultpm{0.2756}{0.01} &\resultpm{0.2395}{0.02} &\resultpm{0.2344}{0.01} &\resultpm{0.2259}{0.01} &\resultpmul{0.2242}{0.00} &\resultpmul{0.2123}{0.01} &\resultpmul{0.1942}{0.00} &\resultpm{0.2247}{0.01} \\
\cmidrule{2-11}
    & \multirow{3}{*}{Linear}
        & MAE &\resultpm{0.1872}{0.00} &\resultpm{0.1952}{0.01} &\resultpm{0.2141}{0.00} &\resultpm{0.2304}{0.01} &\resultpm{0.1388}{0.01} &\resultpm{0.1497}{0.00} &\resultpm{0.1884}{0.01} & \resultpm{0.2233}{0.04}\\
    & & MAPE(\%) &\resultpm{50.544}{0.85} &\resultpm{49.638}{1.98} &\resultpm{42.502}{0.57} &\resultpm{41.292}{0.74} &\resultpm{34.678}{0.91} &\resultpm{34.388}{0.43} &\resultpm{33.828}{0.42} &\resultpm{34.828}{2.22} \\
    & & RMSE &\resultpm{0.2888}{0.01} &\resultpm{0.2823}{0.01} &\resultpm{0.2644}{0.00} &\resultpm{0.2710}{0.01} &\resultpm{0.2430}{0.01} &\resultpm{0.2342}{0.00} &\resultpm{0.2351}{0.01} &\resultpm{0.2707}{0.04} \\
\cmidrule{2-11}
    & \multirow{3}{*}{DTL\cite{DTL}}
        & MAE &\resultpm{0.2078}{0.00} &\resultpm{0.2015}{0.01} &\resultpm{0.1972}{0.00} &\resultpm{0.2005}{0.00} &\resultpm{0.1593}{0.01}&\resultpm{0.1900}{0.01}&\resultpm{0.1685}{0.01}&\resultpm{0.2014}{0.01}\\
    & & MAPE(\%) &\resultpm{50.202}{1.05} &\resultpm{35.672}{1.36}&\resultpm{33.040}{0.90} &\resultpm{31.4900}{0.36} &\resultpm{34.1940}{1.04} &\resultpm{33.9540}{1.26} &\resultpm{27.6220}{1.41} &\resultpm{29.534}{0.89} \\
    & & RMSE &\resultpm{0.2914}{0.01} &\resultpm{0.2560}{0.01} &\resultpm{0.2480}{0.00} &\resultpm{0.2511}{0.00} &\resultpm{0.2376}{0.01} &\resultpm{0.2498}{0.01} &\resultpm{0.2202}{0.01} &\resultpm{0.2551}{0.01} \\
\cmidrule{2-11}
    & \multirow{3}{*}{MKDPINN\cite{MKDPINN}}
        & MAE &\resultpmul{0.1581}{0.03}&\resultpmul{0.1427}{0.01} &\resultpmul{0.1629}{0.01} &\resultpmul{0.1641}{0.01} &\resultpm{0.1241}{0.03}&\resultpmul{0.1382}{0.00}&\resultpmul{0.1383}{0.01}&\resultpmul{0.1557}{0.00}\\
    & & MAPE(\%) &\resultpmul{45.070}{0.80} &\resultpm{36.950}{0.41}&\resultpmul{29.070}{1.66} &\resultpmul{27.320}{1.32} &\resultpm{30.930}{0.45} &\resultpm{29.760}{0.60} &\resultpmul{24.130}{0.72} &\resultpmbf{21.580}{0.41} \\
    & & RMSE &\resultpmul{0.2626}{0.05} &\resultpmul{0.2267}{0.02} &\resultpmbf{0.2076}{0.01}&\resultpmul{0.1981}{0.01}&\resultpm{0.2298}{0.01}&\resultpm{0.2161}{0.00}&\resultpm{0.1997}{0.02}&\resultpmul{0.1915}{0.01}\\
\cmidrule{2-11}
    & \multirow{3}{*}{TACDA\cite{TACDA}}
        & MAE &\resultpm{0.3653}{0.05}&\resultpm{0.5506}{0.03}&\resultpm{0.4029}{0.07}&\resultpm{0.3615}{0.03}&\resultpm{0.2876}{0.02}&\resultpm{0.5355}{0.07}&\resultpm{0.4369}{0.01}&\resultpm{0.3213}{0.03}\\
    & & MAPE(\%) &\resultpm{48.950}{0.77}&\resultpm{64.060}{2.52}&\resultpm{51.380}{1.05}&\resultpm{48.700}{0.72} &\resultpm{40.000}{0.80}&\resultpm{59.490}{1.30}&\resultpm{50.800}{0.33}&\resultpm{42.140}{1.71}\\
    & & RMSE &\resultpm{0.3975}{0.03}&\resultpm{0.5943}{0.05}&\resultpm{0.4390}{0.05}&\resultpm{0.3932}{0.02}&\resultpm{0.3038}{0.04}&\resultpm{0.5667}{0.02}&\resultpm{0.4633}{0.07}&\resultpm{0.3392}{0.04}\\
\midrule

\multirow{12}{*}{\rotatebox[origin=c]{90}{\textbf{Supervised Learning}}} 
    & \multirow{3}{*}{ResNet-18\cite{ResNet-18}}
        & MAE &\resultpm{0.2093}{0.00} &\resultpm{0.2190}{0.03} &\resultpm{0.2067}{0.01} &\resultpm{0.1916}{0.00} &\resultpm{0.1526}{0.00} &\resultpm{0.1792}{0.00} &\resultpm{0.1572}{0.00} &\resultpm{0.1928}{0.01} \\
    & & MAPE(\%) &\resultpm{51.466}{0.72} &\resultpm{43.014}{7.01} &\resultpm{37.330}{1.59} &\resultpm{31.370}{0.27} &\resultpm{34.850}{0.44} &\resultpm{34.838}{0.87} &\resultpm{27.406}{0.62} &\resultpm{28.376}{0.54} \\
    & & RMSE &\resultpm{0.2963}{0.00} &\resultpm{0.2825}{0.04} &\resultpm{0.2588}{0.01} &\resultpm{0.2391}{0.00} &\resultpm{0.2403}{0.00} &\resultpm{0.2460}{0.01} &\resultpm{0.2088}{0.00} &\resultpm{0.2434}{0.01} \\
\cmidrule{2-11}
    & \multirow{3}{*}{DAMCNN\cite{DAMCNN}}
        & MAE &\resultpm{0.1802}{0.00} &\resultpm{0.2192}{0.01} &\resultpm{0.2195}{0.00} &\resultpm{0.2317}{0.01} &\resultpm{0.1254}{0.00} &\resultpm{0.1410}{0.00} &\resultpm{0.1709}{0.00} &\resultpm{0.1995}{0.01} \\
    & & MAPE(\%) &\resultpm{50.080}{0.34} &\resultpm{49.092}{1.35} &\resultpm{46.480}{0.37} &\resultpm{43.034}{1.41} &\resultpm{33.476}{0.10} &\resultpm{33.916}{0.20} &\resultpm{34.720}{0.38} &\resultpm{32.132}{0.81} \\
    & & RMSE &\resultpm{0.2909}{0.00} &\resultpm{0.3132}{0.01} &\resultpm{0.2957}{0.00} &\resultpm{0.3005}{0.01} &\resultpm{0.2443}{0.00} &\resultpm{0.2430}{0.00} &\resultpm{0.2549}{0.01} &\resultpm{0.2751}{0.01} \\
\cmidrule{2-11}
    & \multirow{3}{*}{Dual-Mixer\cite{Dual-Mixer}}
        & MAE &\resultpm{0.2353}{0.00} &\resultpm{0.2831}{0.01} &\resultpm{0.2548}{0.01} &\resultpm{0.2407}{0.02} &\resultpm{0.1828}{0.01} &\resultpm{0.2138}{0.01} &\resultpm{0.2274}{0.01} &\resultpm{0.2350}{0.01} \\
    & & MAPE(\%) &\resultpm{55.010}{0.77} &\resultpm{54.500}{2.11} &\resultpm{45.872}{2.53} &\resultpm{40.376}{3.41} &\resultpm{39.130}{0.81} &\resultpm{40.040}{1.32} &\resultpm{38.172}{1.15} &\resultpm{33.874}{1.08} \\
    & & RMSE &\resultpm{0.3209}{0.01} &\resultpm{0.3590}{0.02} &\resultpm{0.3192}{0.01} &\resultpm{0.2986}{0.02} &\resultpm{0.2684}{0.01} &\resultpm{0.2846}{0.01} &\resultpm{0.2901}{0.01} &\resultpm{0.2922}{0.01} \\
\cmidrule{2-11}
    & \multirow{3}{*}{IMDSSN\cite{IMDSSN}}
        & MAE &\resultpm{0.1934}{0.01} &\resultpm{0.2232}{0.01} &\resultpm{0.2219}{0.01} &\resultpm{0.2269}{0.05} &\resultpm{0.1494}{0.01} &\resultpm{0.1629}{0.01} &\resultpm{0.1830}{0.02} &\resultpm{0.1844}{0.01} \\
    & & MAPE(\%) &\resultpm{51.398}{1.35} &\resultpm{52.036}{2.39} &\resultpm{43.828}{2.82} &\resultpm{40.680}{6.11} &\resultpm{36.300}{1.33} &\resultpm{35.854}{0.58} &\resultpm{33.894}{1.43} &\resultpm{28.278}{0.25} \\
    & & RMSE &\resultpm{0.2936}{0.01} &\resultpm{0.3034}{0.01} &\resultpm{0.2733}{0.00} &\resultpm{0.2697}{0.04} &\resultpm{0.2525}{0.01} &\resultpm{0.2452}{0.00} &\resultpm{0.2354}{0.01} &\resultpm{0.2329}{0.00} \\
\bottomrule
\end{tabular}
}
\label{tab:result cmapss}
\end{table*}

\begin{table*}[h]
\centering
\caption{Comparison of RUL prediction performance on small-sample XJTU-SY Bearing datasets across different methods.}
\resizebox{\linewidth}{!}{
\setlength{\tabcolsep}{3pt}
\begin{tabular}{cl|c|c|c|c|c|c|c|c|c}
\toprule
\multicolumn{2}{c|}{Methods} & Metrics & OP-A-1 & OP-A-2 & OP-B-1 & OP-B-2 & OP-B-3 & OP-C-1 & OP-C-2 & OP-C-3 \\
\midrule
\multirow{18}{*}{\rotatebox[origin=c]{90}{\textbf{Transfer Learning}}} 
    & \multirow{3}{*}{\textbf{PEFT-MuTS}}
        & MAE &\resultpmbf{0.1541}{0.01} &\resultpmbf{0.1536}{0.01} &\resultpmbf{0.1270}{0.01} &\resultpmbf{0.1161}{0.03} &\resultpm{0.1962}{0.04} &\resultpmbf{0.0987}{0.00} &\resultpmbf{0.0990}{0.01} &\resultpm{0.0985}{0.00} \\
    & & SMAPE(\%) &\resultpmul{19.672}{3.88} &\resultpmul{18.215}{2.64} &\resultpmul{18.858}{1.49} &\resultpmbf{12.953}{1.52} &\resultpmul{17.392}{1.88} &\resultpmbf{8.827}{0.29} &\resultpmbf{8.637}{0.24} &\resultpmul{8.551}{0.17} \\
    & & RMSE &\resultpmbf{0.2372}{0.01} &\resultpm{0.2379}{0.01} &\resultpmbf{0.1738}{0.02} &\resultpmul{0.2038}{0.06} &\resultpm{0.3296}{0.07} &\resultpmbf{0.1616}{0.01} &\resultpmul{0.1898}{0.02} &\resultpm{0.1986}{0.01} \\
\cmidrule{2-11}
    & \multirow{3}{*}{Full}
        & MAE &\resultpmul{0.1661}{0.02} &\resultpm{0.1728}{0.02} &\resultpm{0.2021}{0.08} &\resultpm{0.2180}{0.30} &\resultpmbf{0.1121}{0.02} &\resultpmul{0.1088}{0.01} &\resultpm{0.1194}{0.01} &\resultpm{0.1067}{0.05} \\
    & & SMAPE(\%) &\resultpmbf{18.410}{1.55} &\resultpm{20.539}{5.32} &\resultpm{24.982}{8.79} &\resultpm{15.231}{6.47} &\resultpmbf{16.346}{4.92} &\resultpm{9.712}{0.62} &\resultpm{9.964}{1.04} &\resultpm{8.612}{1.32} \\
    & & RMSE &\resultpmul{0.2564}{0.02} &\resultpm{0.2705}{0.03} &\resultpm{0.3069}{0.11} &\resultpm{0.3849}{0.58} &\resultpmbf{0.1607}{0.03} &\resultpmul{0.1989}{0.03} &\resultpm{0.2392}{0.02} &\resultpm{0.2828}{0.22} \\
\cmidrule{2-11}
    & \multirow{3}{*}{Linear}
        & MAE &\resultpm{0.2288}{0.06} &\resultpm{0.1747}{0.03} &\resultpm{0.1943}{0.05} &\resultpmul{0.1168}{0.02} &\resultpm{0.2912}{0.12} &\resultpm{0.1505}{0.03} &\resultpm{0.1165}{0.01} &\resultpm{0.1175}{0.02} \\
    & & SMAPE(\%) &\resultpm{26.249}{4.55} &\resultpm{20.000}{4.77} &\resultpm{23.008}{5.99} &\resultpmul{14.875}{1.78} &\resultpm{22.734}{4.41} &\resultpm{11.366}{1.69} &\resultpm{9.882}{1.05} &\resultpm{9.883}{1.31} \\
    & & RMSE &\resultpm{0.3352}{0.13} &\resultpmul{0.2363}{0.03} &\resultpm{0.2629}{0.06} &\resultpmbf{0.1775}{0.03} &\resultpm{0.4401}{0.19} &\resultpm{0.2204}{0.04} &\resultpmbf{0.1838}{0.02} &\resultpm{0.1895}{0.06} \\
\cmidrule{2-11}
    & \multirow{3}{*}{DTL\cite{DTL}}
        & MAE &\resultpm{0.2331}{0.04}&\resultpm{0.1797}{0.02} &\resultpmul{0.1910}{0.00}&\resultpm{0.3097}{0.21}&\resultpm{0.4033}{0.36}&\resultpm{0.2634}{0.09} &\resultpmul{0.1075}{0.01}&\resultpmbf{0.0918}{0.01}\\
    & & SMAPE(\%) &\resultpm{20.900}{0.81}&\resultpm{18.390}{1.30}&\resultpmbf{16.080}{2.95}&\resultpm{19.498}{5.74} &\resultpm{20.652}{7.44}&\resultpm{15.796}{3.03}&\resultpm{9.934}{0.50}&\resultpmbf{8.498}{0.85} \\
    & & RMSE &\resultpm{0.3482}{0.09}&\resultpm{0.2520}{0.03}&\resultpmul{0.2484}{0.01}&\resultpm{0.5242}{0.39} &\resultpm{0.6854}{0.64} &\resultpm{0.3211}{0.10} &\resultpm{0.1963}{0.03} &\resultpmbf{0.1629}{0.02} \\
\cmidrule{2-11}
    & \multirow{3}{*}{MKDPINN\cite{MKDPINN}}
        & MAE &\resultpm{0.2143}{0.10} &\resultpmul{0.1670}{0.21} &\resultpm{0.1968}{0.11} &\resultpm{0.2593}{0.09} &\resultpm{0.2257}{0.17} &\resultpm{0.1161}{0.15} &\resultpm{0.1269}{0.09} &\resultpm{0.1182}{0.12} \\
    & & SMAPE(\%) &\resultpm{19.930}{1.07} &\resultpmbf{12.810}{0.84} &\resultpm{22.680}{2.01} &\resultpm{20.370}{1.73} &\resultpm{19.120}{2.15} &\resultpmul{9.160}{0.98} &\resultpmul{9.610}{0.64} &\resultpm{9.130}{1.82} \\
    & & RMSE &\resultpm{0.3194}{0.03} &\resultpmbf{0.2063}{0.01} &\resultpm{0.2568}{0.06} &\resultpm{0.4070}{0.01} &\resultpm{0.3583}{0.04} &\resultpm{0.2591}{0.05} &\resultpm{0.2737}{0.01} &\resultpm{0.2803}{0.01} \\
\cmidrule{2-11}
    & \multirow{3}{*}{TACDA\cite{TACDA}}
        & MAE &\resultpm{0.2848}{0.23} &\resultpm{0.2806}{0.34} &\resultpm{0.4418}{0.16} &\resultpm{0.2732}{0.20} &\resultpm{0.2424}{0.20} &\resultpm{0.8250}{0.33} &\resultpm{0.2204}{0.07} &\resultpm{0.1957}{0.11} \\
    & & SMAPE(\%) &\resultpm{23.860}{1.11} &\resultpm{23.580}{0.92} &\resultpm{39.010}{2.65} &\resultpm{22.210}{2.70} &\resultpm{20.080}{0.86} &\resultpm{84.800}{8.13} &\resultpm{15.240}{0.55} &\resultpm{13.670}{0.90} \\
    & & RMSE &\resultpm{0.3384}{0.04} &\resultpm{0.3372}{0.08} &\resultpm{0.4944}{0.10} &\resultpm{0.3322}{0.11} &\resultpm{0.3388}{0.06} &\resultpm{0.8625}{0.62} &\resultpm{0.2616}{0.09} &\resultpm{0.2551}{0.05} \\
\midrule

\multirow{12}{*}{\rotatebox[origin=c]{90}{\textbf{Supervised Learning}}} 
    & \multirow{3}{*}{ResNet-18\cite{ResNet-18}}
        & MAE &\resultpm{0.2061}{0.03} &\resultpm{0.1821}{0.02} &\resultpm{1.1122}{0.76} &\resultpm{0.5743}{0.52} &\resultpm{0.2489}{0.24} &\resultpm{0.1877}{0.07} &\resultpm{0.1468}{0.08} &\resultpmul{0.0937}{0.01} \\
    & & SMAPE(\%) &\resultpm{26.054}{4.72} &\resultpm{22.812}{2.72} &\resultpm{44.166}{18.41} &\resultpm{25.765}{11.88} &\resultpm{21.941}{4.08} &\resultpm{13.450}{3.85} &\resultpm{10.055}{2.40} &\resultpm{8.597}{0.71} \\
    & & RMSE &\resultpm{0.3251}{0.05} &\resultpm{0.2900}{0.03} &\resultpm{1.7024}{1.15} &\resultpm{1.0705}{0.98} &\resultpm{0.3925}{0.43} &\resultpm{0.3585}{0.10} &\resultpm{0.3663}{0.26} &\resultpmul{0.1801}{0.02} \\
\cmidrule{2-11}
    & \multirow{3}{*}{DAMCNN\cite{DAMCNN}}
        & MAE &\resultpm{0.2669}{0.04} &\resultpm{0.2672}{0.02} &\resultpm{0.2224}{0.03} &\resultpm{0.1991}{0.07} &\resultpmul{0.1869}{0.05} &\resultpm{0.1777}{0.06} &\resultpm{0.1204}{0.00} &\resultpm{0.1203}{0.01} \\
    & & SMAPE(\%) &\resultpm{22.115}{1.79} &\resultpm{22.096}{0.98} &\resultpm{21.207}{3.30} &\resultpm{17.555}{3.22} &\resultpm{17.661}{1.76} &\resultpm{13.202}{2.93} &\resultpm{9.328}{0.14} &\resultpm{9.294}{0.31} \\
    & & RMSE &\resultpm{0.3811}{0.05} &\resultpm{0.3855}{0.01} &\resultpm{0.3222}{0.02} &\resultpm{0.3082}{0.11} &\resultpmul{0.2934}{0.09} &\resultpm{0.2554}{0.05} &\resultpm{0.2682}{0.01} &\resultpm{0.2706}{0.01} \\
\cmidrule{2-11}
    & \multirow{3}{*}{Dual-Mixer\cite{Dual-Mixer}}
        & MAE &\resultpm{0.9226}{0.07} &\resultpm{0.8635}{0.20} &\resultpm{0.8010}{0.43} &\resultpm{0.3484}{0.09} &\resultpm{0.4084}{0.15} &\resultpm{1.1707}{0.12} &\resultpm{0.4324}{0.09} &\resultpm{0.3342}{0.02} \\
    & & SMAPE(\%) &\resultpm{41.473}{1.09} &\resultpm{39.988}{4.75} &\resultpm{39.730}{7.10} &\resultpm{25.347}{1.88} &\resultpm{27.259}{3.46} &\resultpm{39.799}{2.38} &\resultpm{21.701}{2.72} &\resultpm{18.700}{0.65} \\
    & & RMSE &\resultpm{1.0493}{0.11} &\resultpm{0.9841}{0.21} &\resultpm{0.9421}{0.43} &\resultpm{0.4406}{0.11} &\resultpm{0.5202}{0.18} &\resultpm{1.3044}{0.11} &\resultpm{0.5308}{0.09} &\resultpm{0.4340}{0.04} \\
\cmidrule{2-11}
    & \multirow{3}{*}{IMDSSN\cite{IMDSSN}}
        & MAE &\resultpm{3.9874}{3.40} &\resultpm{1.6401}{0.73} &\resultpm{5.1916}{2.80} &\resultpm{3.1417}{1.25} &\resultpm{2.1069}{0.80} &\resultpm{1.9114}{0.28} &\resultpm{1.3881}{0.56} &\resultpm{0.8120}{0.31} \\
    & & SMAPE(\%) &\resultpm{66.552}{12.2} &\resultpm{58.360}{7.87} &\resultpm{71.854}{12.5} &\resultpm{64.163}{8.01} &\resultpm{58.784}{5.50} &\resultpm{55.090}{3.30} &\resultpm{47.662}{5.83} &\resultpm{38.066}{5.29} \\
    & & RMSE &\resultpm{5.9707}{5.10} &\resultpm{2.2283}{1.02} &\resultpm{7.1127}{3.64} &\resultpm{4.4466}{1.69} &\resultpm{3.0528}{1.13} &\resultpm{2.9479}{0.59} &\resultpm{1.9058}{0.73} &\resultpm{1.1510}{0.35} \\
\bottomrule
\end{tabular}
}
\label{tab:result bearing}
\end{table*}

\subsubsection{Results and Analysis}
The overall comparative results on all few-shot datasets are presented in Table~\ref{tab:result cmapss} and Table~\ref{tab:result bearing}. 

The results reported in the table suggest that the pre-trained PEFT-MuTS, Linear, Full, and DTL models tend to outperform conventional supervised learning methods across nearly all evaluation metrics. This observation indicates that, even when the pre-training dataset is not directly related to the RUL prediction task, self-supervised pre-training based on learning general temporal representations, rather than task-specific degradation knowledge, can still lead to performance improvements in few-shot RUL prediction.

From the results, the only difference between the Full and ResNet-18 models is that the former leverages FEI pre-training, while the latter uses standard random initialization. Therefore, comparing these two reveals that even without additional fine-tuning structures, full fine-tuning of a pre-trained representation model significantly benefits few-shot RUL prediction, clearly outperforming the randomly initialized ResNet-18. This highlights the inherent value of cross-domain temporal knowledge for data-driven RUL prediction tasks. However, the large number of parameters involved in full fine-tuning (approximately 15M) results in excessive optimization freedom, which may lead to overfitting risks in few-shot scenarios, and thus its performance is not fully unleashed.

In contrast, the Linear approach only fine-tunes the output layer (a single linear regression layer), drastically reducing the number of trainable parameters to 1024. Even with such a limited capacity, it still outperforms several conventional supervised learning methods in most scenarios by leveraging the pre-trained representation model. When compared to the randomly initialized 15M-parameter ResNet-18, the Linear method achieves comparable or even superior performance in some cases, further emphasizing the effectiveness of the pre-trained backbone. However, its extremely low flexibility hinders its ability to adapt to specific tasks, making it noticeably inferior to the full fine-tuning method.

PEFT-MuTS and DTL achieve a more favorable balance between parameter efficiency and predictive performance, significantly improving fine-tuning efficiency. Compared with the Linear strategy, they introduce a modest increase in the number of trainable parameters while substantially enhancing the model’s adaptability in few-shot scenarios. Moreover, PEFT-MuTS incorporates a dedicated multivariate fusion mechanism in its architecture, enabling more effective utilization of multivariate information during fine-tuning. As a result, PEFT-MuTS consistently outperforms DTL with a clear performance margin.

In addition, it can be observed that the meta-learning-based MKDPINN exhibits unstable performance under extreme few-shot conditions, particularly on the FD002-1, FD002-2, FD004-1, OP-B-1, and OP-B-2 datasets. Although its performance significantly surpasses that of conventional supervised learning methods, it still lags behind self-supervised pre-training-based approaches. Domain adaptation methods perform even worse in these settings. The underlying reason is that such methods heavily rely on the availability of data from known similar equipment or operating conditions. When the number of samples is extremely limited, the prior knowledge provided by these datasets becomes insufficient, severely constraining the transfer process. In contrast, self-supervised learning methods directly learn task-agnostic and transferable representations from large-scale datasets. The improved generalization capability of these representations effectively reduces the difficulty of adapting to downstream tasks, without relying exclusively on observable similar equipment or operating conditions. Nevertheless, pre-trained representations alone are not sufficient. The multivariate fusion mechanism specifically designed in PEFT-MuTS ensures more effective information utilization during adaptation, which is particularly beneficial for datasets with a large number of variables, such as C-MAPSS.

\begin{figure}
    \centering
    \begin{subfigure}{0.99\textwidth}
        \includegraphics[width=\linewidth]{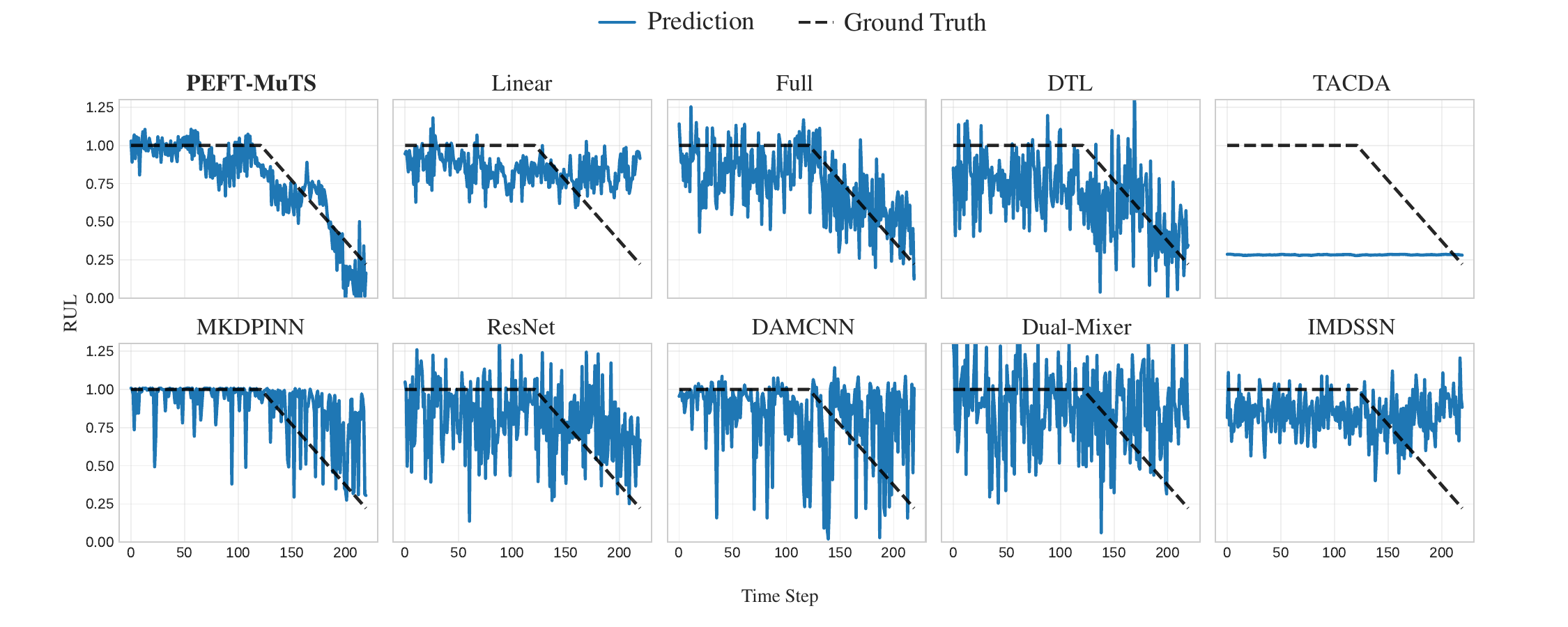}
        \caption{The prediction results of the comparision methods on the 1\# test engine unit of FD002-2 dataset.}
    \label{fig:prediction cmapss}
    \end{subfigure}
    \hfill
    \begin{subfigure}{0.99\textwidth}
        \includegraphics[width=\linewidth]{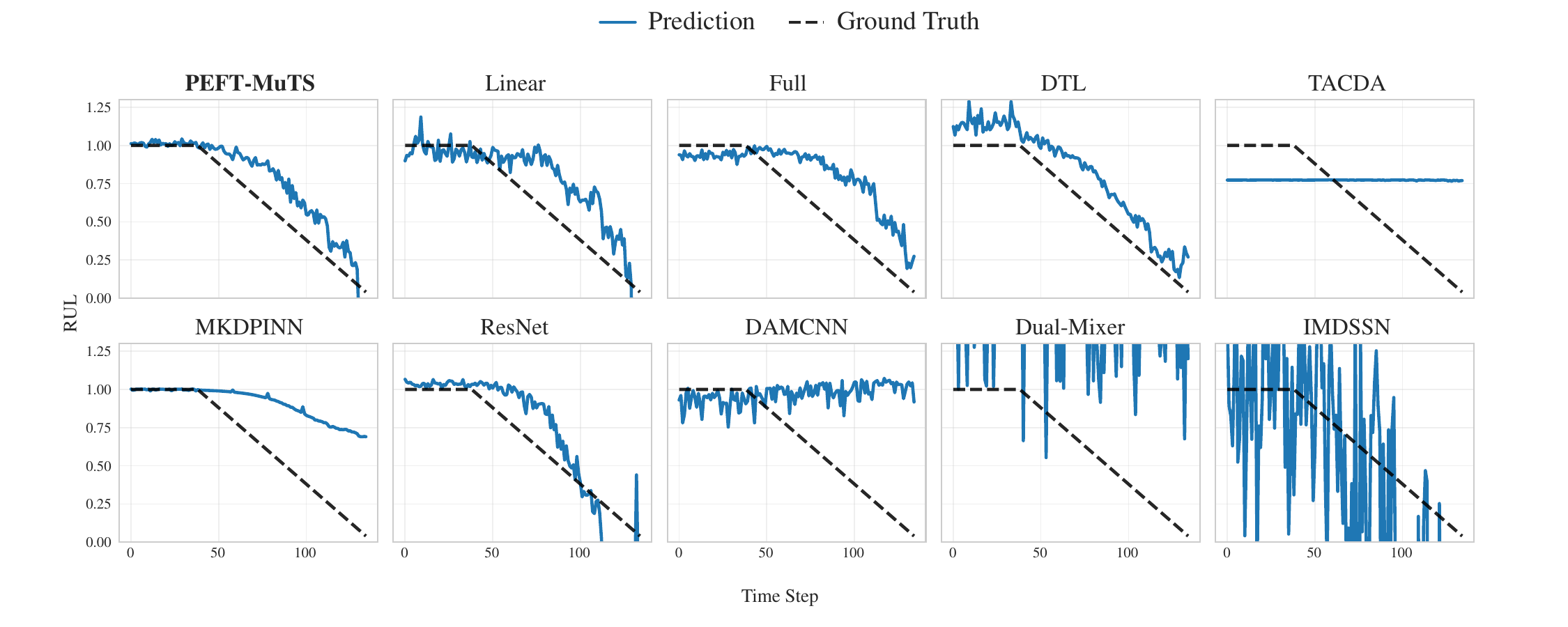}
        \caption{The prediction results of the comparision methods on the 2\# bearing of OP-A-1 dataset.}
    \label{fig:prediction bearing}
    \end{subfigure}
    \caption{The visualization of prediction results on C-MAPSS and Bearing datasets.}
\end{figure}

Figures \ref{fig:prediction cmapss} and \ref{fig:prediction bearing} present representative prediction results of different models on the FD002-2 dataset and the OP-A-1 bearing dataset, respectively. On FD002-2, PEFT-MuTS is able to more accurately model degradation characteristics across the early, middle, and late stages. Due to the scarcity of degradation samples and the predominance of healthy samples, most models exhibit a strong bias toward predicting the healthy stage. Except for the pre-trained Full fine-tuning and DTL methods, which retain limited predictive capability, the remaining models fail to effectively capture degradation trends. On the OP-A-1 dataset, benefiting from the strong generalization ability of the pre-trained model, PEFT-MuTS, Full fine-tuning, DTL, and even Linear fine-tuning achieve more accurate predictions for previously unseen mid-stage degradation samples. In contrast, the randomly initialized ResNet-18 tends to classify all mid- and late-stage degradation samples as late-stage degradation, while DAMCNN shows a strong bias toward early-stage degradation; neither is capable of accurately modeling the mid-stage degradation process. Dual-Mixer and IMDSSN are overly sensitive to noise in vibration signals, resulting in highly fluctuating outputs with little practical predictive value. The domain adaptation method TACDA completely fails under this setting, and the meta-learning-based MKDPINN also delivers unsatisfactory performance.

\begin{table}
    \centering
    \caption{Average training time per epoch of different methods.}
    \begin{tabular}{cccccccccc}
        \toprule
        Models &PEFT-MuTS & Full & Linear & DTL & MKDPINN & TACDA & DAMCNN & Dual-Mixer & IMDSSN \\
        \midrule
        Time (s) &0.089 & 0.101& 0.029& 0.074& 0.096 & 0.122& 0.037 & 0.088 & 0.097 \\
        \bottomrule
    \end{tabular}
    \label{tab:training_time}
\end{table}
Furthermore, we evaluate the computational efficiency of different methods by measuring the average training time per epoch, as reported in Table~\ref{tab:training_time}. Since the Bearing dataset contains substantially longer input sequences than C-MAPSS, the evaluation is conducted on its OP-A-1 subset to provide a more representative comparison of the computational cost. The results show that PEFT-MuTS does not introduce a significant increase in computational complexity compared with the baseline models and maintains a favorable overall computational efficiency.

\subsection{Model Analysis}

\subsubsection{Ablation Study Design}
To evaluate the effectiveness of PEFT-MuTS components, we conduct ablation studies focusing on: (1) cross-domain pre-training of the backbone (\textbf{pre-train}), (2) meta-variable low rank multivariate fusion (\textbf{meta-variable}), and (3) zero-initialized regression head (\textbf{zero-init.}). By selectively enabling or disabling these modules, we construct five variant models.

Specifically, the \textbf{pre-train} setting uses ResNet-18 pre-trained on SleepEEG via the FEI strategy; its removal means using a randomly initialized backbone. The \textbf{meta-variable} module introduces meta-variable $u$ to integrate multivariate features as in Eq.~\ref{eq:u process detail}; without it, features are averaged before output. The \textbf{zero-init.} setting uses a zero-initialized linear regressor; otherwise, a bias-free Kaiming-initialized layer is applied.

\begin{table*}
\centering
\caption{The ablation study results.}
\setlength{\tabcolsep}{3pt}
    \begin{tabular}{ccc|c|ccccc}
         \toprule
         \multicolumn{3}{c|}{Components} & \multirow{2}{*}{Metrics} & \multirow{2}{*}{FD002-1} & \multirow{2}{*}{FD004-1} & \multirow{2}{*}{OP-A-1} & \multirow{2}{*}{OP-B-1} & \multirow{2}{*}{OP-C-1}  \\
         \textbf{pre-train} & \textbf{meta-variable} & \textbf{zero-init.} & & & & & &  \\
         \midrule
         \multirow{3}{*}{\ding{51}} & \multirow{3}{*}{\ding{51}} & \multirow{3}{*}{\ding{51}} & 
         MAE & \resultpmbf{0.1529}{0.01} & \resultpmbf{0.1072}{0.00} & \resultpmbf{0.1541}{0.01} & \resultpm{0.1270}{0.01} & \resultpmbf{0.0987}{0.00} \\
         & & & (S)MAPE(\%) & \resultpmbf{38.382}{2.28} & \resultpmbf{26.582}{0.75} & \resultpm{19.672}{3.88} & \resultpm{18.858}{1.49} & \resultpmbf{8.827}{0.29} \\
         & & & RMSE & \resultpmbf{0.2415}{0.01} & \resultpmbf{0.2082}{0.00} & \resultpmbf{0.2372}{0.02} & \resultpm{0.1738}{0.02} & \resultpmbf{0.1616}{0.01} \\
         \midrule
         \multirow{3}{*}{\ding{51}} & \multirow{3}{*}{\ding{51}} & \multirow{3}{*}{\ding{55}} & 
         MAE & \resultpm{0.1697}{0.01}&\resultpm{0.1243}{0.01} &\resultpm{0.2041}{0.04} &\resultpm{0.1791}{0.06} &\resultpm{0.1396}{0.01} \\
         & & & (S)MAPE(\%) &\resultpm{42.806}{1.52} &\resultpm{29.646}{1.75} &\resultpm{24.538}{5.50} &\resultpm{23.894}{8.20} &\resultpm{10.738}{0.49} \\
         & & & RMSE &\resultpm{0.2531}{0.01} &\resultpm{0.2156}{0.01} &\resultpm{0.3030}{0.06} &\resultpm{0.2433}{0.08} &\resultpm{0.2163}{0.03} \\
         \midrule
         \multirow{3}{*}{\ding{51}} & \multirow{3}{*}{\ding{55}} & \multirow{3}{*}{\ding{51}} & 
         MAE &\resultpm{0.2081}{0.00} &\resultpm{0.1796}{0.01} &\resultpm{0.2135}{0.01} &\resultpmbf{0.1025}{0.01} &\resultpm{0.1325}{0.01} \\
         & & & (S)MAPE(\%) &\resultpm{49.938}{0.72} &\resultpm{36.296}{1.02} &\resultpmbf{19.150}{0.82} &\resultpmbf{14.070}{0.44} &\resultpm{10.228}{0.27} \\
         & & & RMSE &\resultpm{0.2899}{0.00} &\resultpm{0.2525}{0.01} &\resultpm{0.2693}{0.01} &\resultpmbf{0.1367}{0.01} &\resultpm{0.1921}{0.01} \\
         \midrule
         \multirow{3}{*}{\ding{55}} & \multirow{3}{*}{\ding{51}} & \multirow{3}{*}{\ding{51}} & 
         MAE &\resultpm{0.1662}{0.01} &\resultpm{0.1138}{0.01} &\resultpm{0.2691}{0.01} &\resultpm{0.1737}{0.04} &\resultpm{0.1130}{0.01} \\
         & & & (S)MAPE(\%) &\resultpm{43.518}{2.72} &\resultpm{28.090}{1.96} &\resultpm{28.484}{0.90} &\resultpm{22.438}{3.39} &\resultpm{9.608}{0.30} \\
         & & & RMSE &\resultpm{0.2609}{0.01} &\resultpm{0.2132}{0.01} &\resultpm{0.4574}{0.02} &\resultpm{0.2799}{0.06} &\resultpm{0.2431}{0.01} \\
         \midrule
         \multirow{3}{*}{\ding{51}} & \multirow{3}{*}{\ding{55}} & \multirow{3}{*}{\ding{55}} & 
         MAE &\resultpm{0.2059}{0.00} &\resultpm{0.1647}{0.01} &\resultpm{0.2420}{0.07} &\resultpm{0.1829}{0.03} &\resultpm{0.1737}{0.03} \\
         & & & (S)MAPE(\%) &\resultpm{50.194}{0.84} &\resultpm{34.990}{1.27} &\resultpm{21.634}{2.35} &\resultpm{19.832}{4.82} &\resultpm{11.826}{1.34} \\
         & & & RMSE &\resultpm{0.2900}{0.00} &\resultpm{0.2419}{0.01} &\resultpm{0.3330}{0.07} &\resultpm{0.2589}{0.03} &\resultpm{0.2541}{0.04} \\
         \midrule
         \multirow{3}{*}{\ding{55}} & \multirow{3}{*}{\ding{51}} & \multirow{3}{*}{\ding{55}} & 
         MAE &\resultpm{0.1931}{0.00} &\resultpm{0.1451}{0.00} &\resultpm{0.3338}{0.05} &\resultpm{0.2198}{0.05} &\resultpm{0.1428}{0.03} \\
         & & & (S)MAPE(\%) &\resultpm{49.064}{0.97} &\resultpm{33.790}{0.44} &\resultpm{30.714}{3.11} &\resultpm{19.882}{2.03} &\resultpm{10.408}{0.92} \\
         & & & RMSE &\resultpm{0.2859}{0.01} &\resultpm{0.2383}{0.00} &\resultpm{0.5526}{0.11} &\resultpm{0.3144}{0.06} &\resultpm{0.2884}{0.05} \\
         \midrule
         \multirow{3}{*}{\ding{55}} & \multirow{3}{*}{\ding{55}} & \multirow{3}{*}{\ding{51}} & 
         MAE &\resultpm{0.1969}{0.00} &\resultpm{0.1594}{0.00} &\resultpm{0.2805}{0.05} &\resultpm{0.1533}{0.10} &\resultpm{0.1170}{0.02} \\
         & & & (S)MAPE(\%) &\resultpm{50.042}{0.46} &\resultpm{35.146}{0.38} &\resultpm{28.466}{4.56} &\resultpm{17.072}{4.35} &\resultpm{9.690}{0.70} \\
         & & & RMSE &\resultpm{0.2865}{0.00} &\resultpm{0.2414}{0.00} &\resultpm{0.4358}{0.08} &\resultpm{0.2501}{0.13} &\resultpm{0.2429}{0.02} \\
         \midrule
         \multirow{3}{*}{\ding{55}} & \multirow{3}{*}{\ding{55}} & \multirow{3}{*}{\ding{55}} & 
         MAE &\resultpm{0.2055}{0.00} &\resultpm{0.1654}{0.00} &\resultpm{0.5510}{0.19} &\resultpm{0.2750}{0.14} &\resultpm{0.1724}{0.12} \\
         & & & (S)MAPE(\%) &\resultpm{50.774}{0.55} &\resultpm{35.880}{0.39} &\resultpm{35.366}{4.46} &\resultpm{22.500}{3.80} &\resultpm{10.986}{3.25} \\
         & & & RMSE &\resultpm{0.2919}{0.00} &\resultpm{0.2466}{0.00} &\resultpm{0.9542}{0.37} &\resultpm{0.4336}{0.23} &\resultpm{0.3458}{0.22} \\
         \bottomrule
    \end{tabular}
    \label{tab:ablation}
\end{table*}

\subsubsection{Ablation Results and Analysis}
The ablation results are shown in Table~\ref{tab:ablation}. Several key findings emerge:

\begin{enumerate}
    \item \textbf{Cross-domain pre-training:}  The cross-domain pre-training of the backbone significantly improves performance on the few-shot RUL prediction task. Removing this component leads to notable degradation across all datasets. However, its effectiveness depends on complementary designs that account for multivariate fusion and training stability (i.e., zero-initialized regression). The meta-variable low rank fusion and zero-initialized regressor proposed in this work synergize with pre-training and together yield substantial improvements under extreme few-shot conditions.
    \item \textbf{Multivariate fusion:}  The meta-variable-based low rank fusion method proposed in PEFT-MuTS consistently improves prediction performance across almost all datasets. On the Bearing dataset, the improvement is relatively limited, which is attributed to the small number of variables and the strong inter-variable correlations--where single-variable features already provide sufficient information for modeling degradation. This suggests that RUL prediction methods based on pre-trained models should consider the nature of the multivariate context. When variable count is low and multivariate information is less critical, a combination of cross-domain pre-trained backbones and zero-initialized regression suffices. Our meta-variable low rank multivariate fusion method is nonetheless applicable to most practical scenarios where multivariate modeling is essential.
    \item \textbf{Zero-initialization:}  Zero-initialization provides a simple yet effective strategy for stabilizing training in pre-trained RUL prediction models. Models with zero-initialized regressors exhibit significantly lower variance across multiple runs, especially for MAPE and SMAPE metrics. Notably, even without backbone pre-training, the zero-initialized regressor greatly reduces prediction errors and can serve as a general-purpose regression head.
\end{enumerate}
\subsubsection{Gate Mechanism Validation}
To further validate the effectiveness of the proposed Gate Mechanism for meta-variable-based multivariate fusion, we compare it with two alternative fusion strategies: an Attention-based fusion method and a Mean Pooling fusion method. In the Attention-based method, the gating module is replaced with a multi-head self-attention layer (2 attention heads), while all subsequent operations remain unchanged. In the Mean Pooling method, the gating module is removed, and the meta-variable is obtained by directly averaging all input variables. The experimental results are summarized in Table~\ref{tab:fusion_comparison}.

As shown in Table~\ref{tab:fusion_comparison}, the proposed Gate Mechanism consistently achieves the best performance across all datasets and evaluation metrics. Although self-attention has been widely adopted for multivariate feature fusion, its relatively higher model complexity requires more training data to learn reliable attention weights. Consequently, under the few-shot setting, the Attention-based fusion method performs noticeably worse than the proposed Gate Mechanism and, in most cases, even underperforms the simple Mean Pooling strategy.

In contrast, the Mean Pooling method treats all variables equally and lacks the flexibility to adaptively adjust the contribution of individual variables, resulting in inferior performance compared with the proposed Gate Mechanism. Overall, the Gate Mechanism achieves a favorable balance between model complexity and adaptive fusion capability, making it more suitable for few-shot multivariate RUL prediction.
\begin{table}
    \centering
    \caption{The performance of Attention fusion, Mean Pool fusion compared with proposed Gate mechanism in meta-variable.}
    \begin{tabular}{c|c|ccccc}
         \toprule
         Method & Metrics & FD002-1 & FD004-1 & OP-A-1 & OP-B-1 & OP-C-1  \\ 
         \midrule
         \multirow{3}{*}{\textbf{Gate}} & MAE & \resultpmbf{0.1529}{0.01} & \resultpmbf{0.1072}{0.00} & \resultpmbf{0.1541}{0.01} & \resultpmbf{0.1270}{0.01} & \resultpmbf{0.0987}{0.00}  \\
         & (S)MAPE(\%) & \resultpmbf{38.382}{2.28} & \resultpmbf{26.582}{0.75} & \resultpmbf{19.672}{3.88} & \resultpmbf{18.858}{1.49} & \resultpmbf{8.827}{0.29}\\
         & RMSE & \resultpmbf{0.2415}{0.01} & \resultpmbf{0.2082}{0.00} & \resultpmbf{0.2372}{0.02} & \resultpmbf{0.1738}{0.02} & \resultpmbf{0.1616}{0.01} \\
         \midrule
         \multirow{3}{*}{Attention} & MAE &\resultpm{0.1660}{0.00} &\resultpm{0.1189}{0.00} &\resultpm{0.1967}{0.02} &\resultpm{0.1319}{0.00} &\resultpm{0.1897}{0.04} \\
         & (S)MAPE(\%) &\resultpm{43.564}{1.12} &\resultpm{30.366}{1.32} &\resultpm{29.150}{2.50} &\resultpm{19.596}{0.48} &\resultpm{12.692}{1.47} \\
         & RMSE &\resultpm{0.2641}{0.01} &\resultpm{0.2270}{0.01} &\resultpm{0.3032}{0.03} &\resultpm{0.2495}{0.01} &\resultpm{0.2439}{0.04} \\
         \midrule
         \multirow{3}{*}{Mean} & MAE &\resultpm{0.1595}{0.01} &\resultpm{0.1136}{0.01} &\resultpm{0.2015}{0.01} &\resultpm{0.1350}{0.02} &\resultpm{0.1583}{0.02} \\
         & (S)MAPE(\%) &\resultpm{40.984}{2.53} &\resultpm{28.680}{1.99} &\resultpm{19.882}{1.04} &\resultpm{19.288}{0.52} &\resultpm{11.278}{0.86} \\
         & RMSE &\resultpm{0.2527}{0.01} &\resultpm{0.2181}{0.01} &\resultpm{0.3277}{0.01} &\resultpm{0.2299}{0.05} &\resultpm{0.2084}{0.02} \\
         \bottomrule
    \end{tabular}
    \label{tab:fusion_comparison}
\end{table}


\subsubsection{Fine-tuning Stability}
To further validate the effectiveness of zero-initialization and its impact on the optimization process, we conduct stability experiments. We compare 3 configurations: ResNet-18 pre-trained with FEI (marked as \textbf{Pre-trained}), ResNet-18 with the full PEFT-MuTS architecture (marked as \textbf{PEFT-MuTS}), and randomly initialized ResNet-18 (marked as \textbf{Random Init.}).

\begin{figure}
    \centering
    \includegraphics[width=0.8\linewidth]{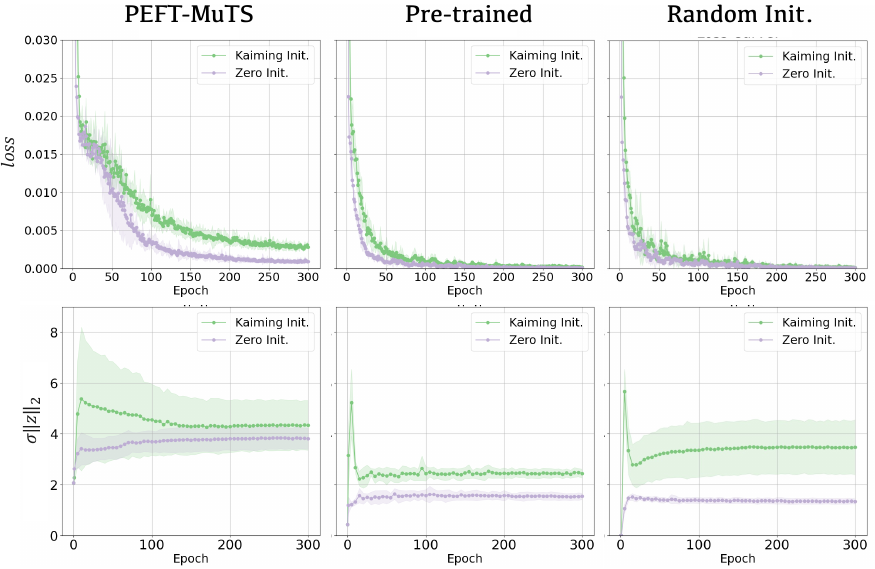}
    \caption{Evolution of training loss and the std of feature $\ell_2$ norms over training epochs.}
    \label{fig:training_variance}
\end{figure}

We fix a set of evaluation samples that are excluded from the gradient update process and record the standard deviation(std) of their backbone output feature's $\ell_2$ norm across batches, denoted as $\sigma ||z||_2$. A larger variance indicates a more unstable training process, which is also reflected in the variance of the training loss curves. We visualize the training process on the FD002-1 dataset, as shown in Figure~\ref{fig:training_variance}. This includes both the training loss variance curves and the $\sigma ||z||_2$ trajectories for the evaluation samples.

In the figure, green curves represent results using Kaiming-initialized regressors, while purple curves correspond to zero-initialized ones. Shaded regions show the standard deviation across multiple runs. We observe the following:

All models exhibit an initial fluctuation phase (first 10 epochs), during which the feature std $\sigma ||z||_2$ first increases then stabilizes. High early std leads to greater divergence among runs, which is particularly pronounced in the PEFT-MuTS loss curves--consistent with the theoretical analysis in Section~\ref{section: zero-initialized}.  
After applying zero-initialization (purple curves), the variance from \eqref{eq:unstable analysis} is suppressed. The std of the feature norm stabilizes more quickly, reducing loss value and accelerating convergence toward lower loss values in PEFT-MuTS. The positive effect of zero-initialization is more prominent for pre-trained models (PEFT-MuTS, Pre-trained) than for randomly initialized ones. This is because pre-training yields high initial feature variance, while random initialization naturally leads to low variance, again aligning with the theoretical analysis.

Combined with the ablation results in Table~\ref{tab:ablation}, we conclude that the zero-initialized regressor greatly enhances convergence speed and generalization when used in pre-trained-based RUL prediction models.

\subsubsection{Different Pre-training Methods}
To further verify that the performance gain of cross-domain pre-training is not specific to the FEI framework, we evaluate two additional representative self-supervised time-series pre-training methods, SimMTM\cite{SimMTM} and InfoTS\cite{InfoTSNew}. For each pretrained backbone, we further compare two regressor initialization strategies: Kaiming initialization and the proposed zero-initialization. The experimental results are summarized in Table~\ref{tab:different_pretrain_result}. For a fair comparison, all methods employ the same ResNet-18 backbone.

As shown in Table~\ref{tab:different_pretrain_result}, all pretrained models consistently outperform the corresponding ResNet-18 model trained from scratch (see Tables \ref{tab:result cmapss} and \ref{tab:result bearing}), demonstrating that the performance gain brought by cross-domain self-supervised time-series pre-training is not limited to a specific pre-training method, but is generally applicable to few-shot RUL prediction. 

Furthermore, the proposed zero-initialization consistently achieves better performance than conventional Kaiming initialization in the vast majority of cases across all three pre-training methods. This indicates that the proposed initialization strategy is not tied to a particular pretrained model, but serves as a generally effective fine-tuning strategy for cross-domain pretrained RUL prediction models.
\begin{table}
    \centering
    \caption{Performance comparison of different self-supervised pre-training methods and regressor initialization strategies.}
    \begin{tabular}{cc|c|ccccc}
        \toprule
        Pre-training & Regressor & Metrics & FD002-1 &FD004-1 &OP-A-1 &OP-B-1 & OP-C-1  \\
        \midrule
        \multirow{6}{*}{FEI} & \multirow{3}{*}{Kaiming init.} & MAE & \resultpm{0.1697}{0.01}&\resultpm{0.1243}{0.01} &\resultpmul{0.2041}{0.04} &\resultpm{0.1791}{0.06} &\resultpm{0.1396}{0.01} \\
          & & (S)MAPE &\resultpm{42.806}{1.52} &\resultpm{29.646}{1.75} &\resultpm{24.538}{5.50} &\resultpmul{23.894}{8.20} &\resultpm{10.738}{0.49} \\
          & & RMSE &\resultpm{0.2531}{0.01} &\resultpm{0.2156}{0.01} &\resultpmul{0.3030}{0.06} &\resultpmul{0.2433}{0.08} &\resultpmul{0.2163}{0.03} \\
        \cmidrule{2-8}
          & \multirow{3}{*}{zero-init.} & MAE & \resultpmbf{0.1529}{0.01} & \resultpmbf{0.1072}{0.00} & \resultpmbf{0.1541}{0.01} & \resultpmbf{0.1270}{0.01} & \resultpmbf{0.0987}{0.00} \\
          & & (S)MAPE  & \resultpmbf{38.382}{2.28} & \resultpmbf{26.582}{0.75} & \resultpmbf{19.672}{3.88} & \resultpmbf{18.858}{1.49} & \resultpmbf{8.827}{0.29} \\
          & & RMSE & \resultpmbf{0.2415}{0.01} & \resultpmbf{0.2082}{0.00} & \resultpmbf{0.2372}{0.02} & \resultpmbf{0.1738}{0.02} & \resultpmbf{0.1616}{0.01} \\
        \midrule
        \multirow{6}{*}{SimMTM} & \multirow{3}{*}{Kaiming init.} & MAE &\resultpm{0.1636}{0.00} &\resultpm{0.1170}{0.00} &\resultpm{0.2818}{0.09} &\resultpm{0.2972}{0.15} &\resultpmul{0.1280}{0.02} \\
          & & (S)MAPE &\resultpm{41.544}{0.89} &\resultpm{27.972}{0.32} &\resultpm{22.012}{3.02} &\resultpm{30.392}{13.20} &\resultpm{11.326}{2.49} \\
          & & RMSE &\resultpm{0.2524}{0.00} &\resultpm{0.2124}{0.00} &\resultpm{0.5522}{0.28} &\resultpm{0.4327}{0.20} &\resultpm{0.2270}{0.08} \\
        \cmidrule{2-8}
          & \multirow{3}{*}{zero-init.} & MAE &\resultpm{0.1585}{0.01} &\resultpm{0.1133}{0.00} &\resultpm{0.3126}{0.05} &\resultpm{0.2486}{0.05} &\resultpm{0.1364}{0.02} \\
          & & (S)MAPE &\resultpmul{40.718}{2.40} &\resultpm{27.858}{1.11} &\resultpm{21.778}{1.41} &\resultpm{28.684}{2.95} &\resultpm{11.994}{2.05} \\
          & & RMSE &\resultpmul{0.2484}{0.01} &\resultpm{0.2080}{0.01} &\resultpm{0.8790}{0.09} &\resultpm{0.3534}{0.07} &\resultpm{0.2253}{0.01} \\
        \midrule
        \multirow{6}{*}{InfoTS} & \multirow{3}{*}{Kaiming init.} & MAE &\resultpm{0.1643}{0.01} &\resultpm{0.1151}{0.00} &\resultpm{0.5269}{0.57} &\resultpm{1.1120}{1.15} &\resultpm{0.2723}{0.18} \\
          & & (S)MAPE &\resultpm{42.476}{1.52} &\resultpm{28.072}{1.15} &\resultpm{22.534}{3.38} &\resultpm{29.466}{5.85} &\resultpm{10.548}{1.30} \\
          & & RMSE &\resultpm{0.2571}{0.01} &\resultpm{0.2136}{0.01} &\resultpm{1.4642}{2.04} &\resultpm{3.2132}{3.59} &\resultpm{1.6112}{1.57} \\
        \cmidrule{2-8}
          & \multirow{3}{*}{zero-init.} & MAE &\resultpmul{0.1632}{0.00} &\resultpmul{0.1124}{0.00} &\resultpm{0.5005}{0.13} &\resultpm{0.5737}{0.42} &\resultpm{0.2398}{0.17} \\
          & & (S)MAPE &\resultpm{41.796}{1.08} &\resultpmul{27.516}{0.84} &\resultpmul{21.016}{3.08} &\resultpm{29.022}{6.29} &\resultpmul{10.518}{1.33} \\
          & & RMSE &\resultpmul{0.2484}{0.01} &\resultpmul{0.2116}{0.00} &\resultpm{1.2488}{0.40} &\resultpm{1.3713}{1.24} &\resultpm{1.2309}{1.72} \\
        \bottomrule
    \end{tabular}
    \label{tab:different_pretrain_result}
\end{table}

\subsubsection{Hyperparameter Sensitivity}
\begin{figure}
    \centering
    \includegraphics[width=0.8\linewidth]{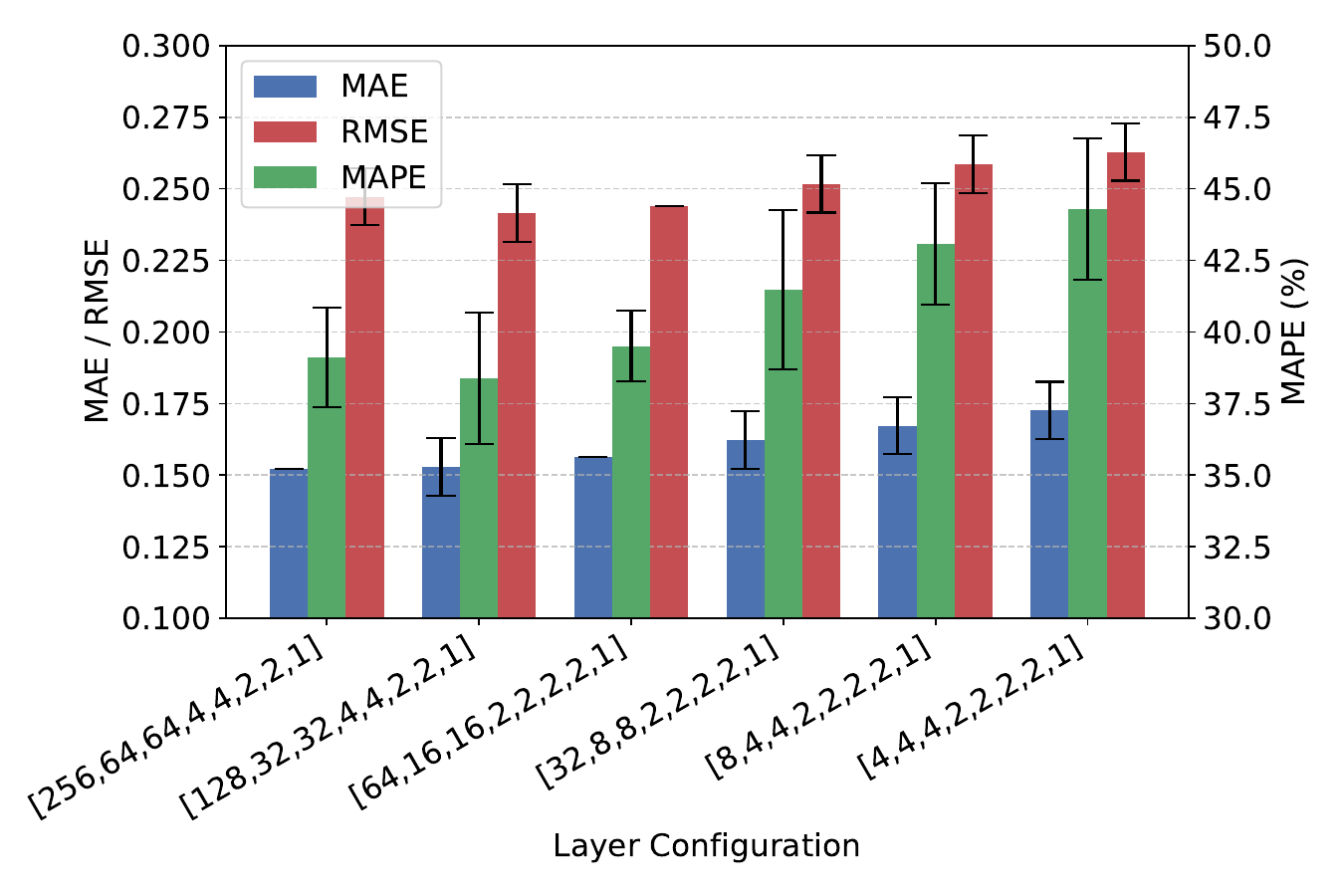}
    \caption{Performance under different projected dimensions $r$ on the FD002-1 dataset.}
    \label{fig:r_config}
\end{figure}
In Side Tuning-based PEFT methods, the projected dimension $r_k$ (i.e., of $\mathbf{A}_k$ and $\mathbf{B}_k$ in \eqref{eq:xi process overall}) is a key hyperparameter. Contrary to the convention in CV and NLP that favors extremely low $r_k$ values\cite{PEFT, DTL}, our experiments show that allocating larger $r_k$ to shallow layers and smaller $r_k$ to deeper ones yields better performance under a fixed parameter budget (see Figure~\ref{fig:r_config}).

The horizontal axis in Figure~\ref{fig:r_config} shows different layer-wise $r_k$ settings (e.g., [256, 64, 64, 4, 4, 2, 2, 1] from shallow to deep). Increasing $r_k$ in early layers consistently reduces test error. However, excessively large $r_k$ can inflate parameter count--for example, 88.5K in the first configuration vs. 52.9K in the second--potentially leading to overfitting.

Given that ResNet-18 generates high-dimensional features in deeper layers, using large $r_k$ there would cause exponential parameter growth. To balance efficiency and effectiveness, we adopt smaller $r_k$ values in deeper layers, aligning with standard practices in PEFT\cite{Side_Tuning_1}\cite{Side_Tuning_2}\cite{DTL}.

\section{Conclusion and Future Work}
This paper proposes PEFT-MuTS, a novel framework to integrate cross-domain representation learning with PEFT for few-shot RUL prediction. By leveraging cross-domain pre-training, PEFT-MuTS reduces reliance on scarce in-domain data and generalizes well with minimal degradation samples. The proposed PEFT-MuTS framework successfully adapts univariate pretrained backbone networks to multivariate RUL prediction tasks through the Independent Feature Tuning Network and Meta-Variable–based Low-Rank Feature Fusion. By leveraging a simple yet effective zero-initialized regressor, it significantly improves fine-tuning stability when applying pretrained models. On the C-MAPSS and XJTU-SY Bearing datasets, PEFT-MuTS substantially outperforms existing few-shot RUL prediction methods using less than 1\% of the available data, demonstrating the effectiveness of cross-domain temporal knowledge transfer for RUL prediction. This framework extends the transfer perspective from “identifying similar degradation patterns” to “leveraging generalizable temporal priors,” offering a new paradigm for few-shot RUL prediction research. 

Potential directions for future research include:
\begin{enumerate}
    \item Source-domain transferability: Investigating how to quantify the transferability of source-domain data to target-domain degradation sequences, thereby enhancing the interpretability of general temporal knowledge for degradation time series.
    \item Hybrid sample utilization: Combining cross-domain samples with domain-specific samples from similar equipment to improve few-shot prediction accuracy when data from similar devices are available.
    \item Practical industrial scenarios: Real-world industrial applications often involve additional challenges, such as noisy RUL labels, missing or faulty sensor measurements, variable operating conditions, censored degradation trajectories, and highly nonlinear degradation behaviors. Extending cross-domain pre-training to improve its robustness under these practical conditions remains an important direction for future research.
\end{enumerate}


\clearpage

\printcredits

\bibliographystyle{unsrt}

\bibliography{cas-refs}






\end{document}